\documentclass[letterpaper,8pt]{extarticle}  
\usepackage{import}
\usepackage{local}
\usepackage[left=1.00in,top=1.00in,bottom=1.00in,right=1.00in]{geometry}
\usepackage{lipsum} 

\usepackage{hyperref}
\usepackage{breakurl}
\usepackage{tabularx}

\usepackage{svg}  

\title{D4: Text-guided diffusion model-based domain adaptive data augmentation for vineyard shoot detection} 

\author{%
    \textbf{
        Kentaro Hirahara \textcolor{Accent}{\textsuperscript{1}}, %
        Chikahito Nakane \textcolor{Accent}{\textsuperscript{2}}, %
        Hajime Ebisawa   \textcolor{Accent}{\textsuperscript{2}}, %
        Tsuyoshi Kuroda  \textcolor{Accent}{\textsuperscript{3}}, %
        Yohei Iwaki      \textcolor{Accent}{\textsuperscript{3}}, %
        Tomoyoshi Utsumi \textcolor{Accent}{\textsuperscript{3}}, %
        Yuichiro Nomura  \textcolor{Accent}{\textsuperscript{4}}, %
        Makoto Koike     \textcolor{Accent}{\textsuperscript{5}}, %
        Hiroshi Mineno   \textcolor{Accent}{\textsuperscript{1,2,4,5,*}
    } } \leavevmode \\ \leavevmode \\
    \begin{small}
        \textcolor{Accent}{\textsuperscript{1}} Graduate School of Integrated Science and Technology, Shizuoka University, 3-5-1 Johoku, Chuo-ku, Hamamatsu, Shizuoka 432-8011, Japan \\ 
        \textcolor{Accent}{\textsuperscript{2}} Faculty of Informatics, Shizuoka University, 3-5-1 Johoku, Chuo-ku, Hamamatsu, Shizuoka 432-8011, Japan \\
        \textcolor{Accent}{\textsuperscript{3}} Technical Research \& Development Center, Yamaha Motor Co., Ltd., 2500 Shingai, Iwata, Shizuoka 438-8501, Japan \\
        \textcolor{Accent}{\textsuperscript{4}} Research Institute of Green Science and Technology, Shizuoka University, 836 Ohya, Shizuoka City, Shizuoka, 422-8529, Japan \\
        \textcolor{Accent}{\textsuperscript{5}} Graduate School of Science and Technology, Shizuoka University, 3-5-1 Johoku, Chuo-ku, Hamamatsu, Shizuoka 432-8011, Japan \\
        \textcolor{Accent}{\textsuperscript{*}} corresponding author at: 3-5-1 Johoku, Chuo-ku, Hamamatsu, Shizuoka 432-8011, Japan: \textcolor{Accent}{mineno@inf.shizuoka.ac.jp} \\ 
    \end{small}
}
\date{}
\begin{document}
\maketitle

\section*{Abstract}
    In an agricultural field, plant phenotyping using object detection models is gaining attention. 
    Plant phenotyping is a technology that accurately measures the quality and condition of cultivated crops from images, contributing to the improvement of crop yield and quality, as well as reducing environmental impact. 
    However, collecting the training data necessary to create generic and high-precision models is extremely challenging due to the difficulty of annotation and the diversity of domains.
    This difficulty arises from the unique shapes and backgrounds of plants, as well as the significant changes in appearance due to environmental conditions and growth stages. 
    Furthermore, it is difficult to transfer training data across different crops, and although machine learning models effective for specific environments, conditions, or crops have been developed, they cannot be widely applied in actual fields. 
    We faced such challenges in the shoot detection task in vineyard.
    Therefore, in this study, we propose a generative artificial intelligence data augmentation method (D4). 
    D4 uses a pre-trained text-guided diffusion model based on a large number of original images culled from video data collected by unmanned ground vehicles or other means, and a small number of annotated datasets.
    The proposed method generates new annotated images with background information adapted to the target domain while retaining annotation information necessary for object detection. 
    In addition, D4 overcomes the lack of training data in agriculture, including the difficulty of annotation and diversity of domains. 
    We confirmed that this generative data augmentation method improved the mean average precision by up to 28.65\% for the BBox detection task and the average precision by up to 13.73\% for the keypoint detection task for vineyard shoot detection. 
    D4 generative data augmentation is expected to simultaneously solve the cost and domain diversity issues of training data generation in agriculture and improve the generalization performance of detection models.
    
\leavevmode \\ \leavevmode \\

\section*{Keywords}
    Agriculture; generative data augmentation; domain adaptation; image-based phenotyping; image generation 

\newpage
\section{Introduction} 
    Precision agriculture (PA) plays a crucial role in sustainable food production systems, and it involves a detailed recording of the conditions of farmland and crops and the precise control of the cultivation environment, which contributes to improving crop yield and quality and reducing environmental impact \citep{Precision_Agriculture_1}\citep{Precision_Agriculture_2}\citep{Precision_Agriculture_3}. 
    Thus, PA enables us to optimize land use and reduce pesticide and fertilizer usage, thereby enhancing productivity while reducing the environmental load. 
    Plant phenotyping technology accurately measures the growth status of crops, and therefore, realizing PA is essential for this technology. 

    Object detection models based on deep learning have recently attracted considerable research attention \citep{qiao_editorial_2022}. 
    When object detection models are used for plant phenotyping, the quality and condition of crops can be easily quantified by detecting fruits and other parts from the acquired images and by performing fruit counting \citep{rahim_deep_2022}\citep{afonso_tomato_2020} and growth analysis \citep{cho_plant_2023} for yield prediction.
    In addition, accumulating the growth process as time-series data and providing cultivation feedback based on comparisons with past data can help optimize environmental  control \citep{wakamori_multimodal_2020}.

\begin{figure}[H] 
            \begin{subcaptiongroup}
                \phantomcaption\label{fig:DifficultAnnotation_a}
                \phantomcaption\label{fig:DifficultAnnotation_b}
                \phantomcaption\label{fig:DifficultAnnotation_c}
            \end{subcaptiongroup}
            \includegraphics[width=0.7\textwidth]{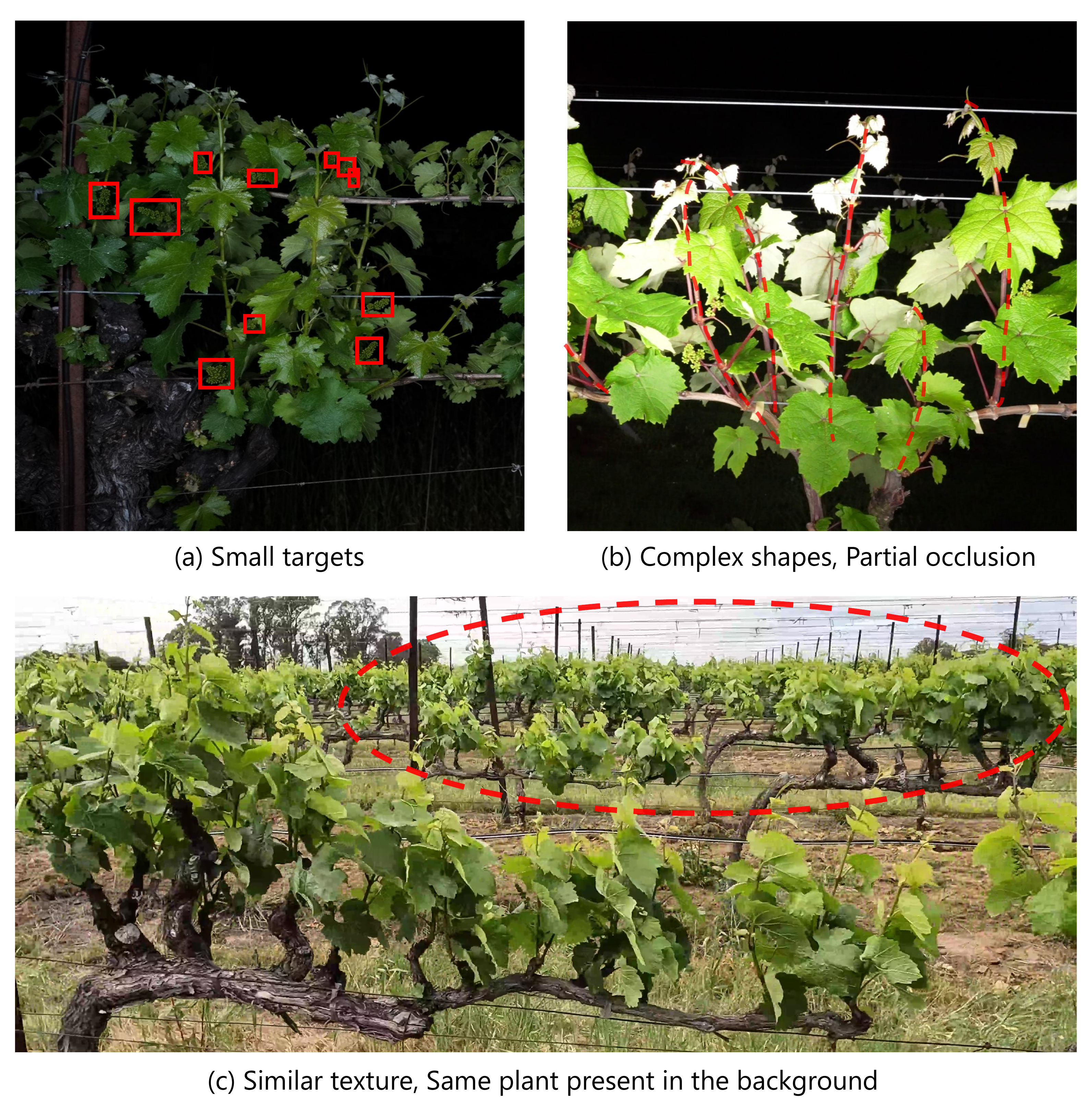}
            \caption{\textcolor{Highlight}{\textbf{Challenges in annotation tasks for vineyard cultivation}}\\
            (a) Detection and counting of very small inflorescences\\
            (b) Shape complexity and partial occlusion in shoot detection\\
            (c) Influence of background vegetation and similar textures in images taken during daylight
            }
            \label{fig:DifficultAnnotation}
        \end{figure}

    Despite the advantages in plant phenotyping, object detection models cannot be used for obtaining object-annotated images for model training. 
    Securing such object-annotated images requires images with annotations, such as bounding box (BBox) and keypoint, making this data collection process time consuming and costly.
    For example, the COCO dataset \citep{lin_microsoft_2014} used for general object recognition tasks such as people and cars contains approximately 328,000 images and 1.5 million BBox annotations.
    In contrast, the GWHD dataset \citep{david_global_2021} for wheat head detection in an agricultural field provides 6,422 images and 275,167 BBox annotations.
    This comparison shows that the number of images and annotations is limited compared to those of other typical datasets. 
    The lack of training data in agriculture is related to the ``difficulty of annotation'' and the ``diversity of domains.''
    \Cref{fig:DifficultAnnotation} shows an example of the annotations for object detection in vine grape cultivation aimed at plant phenotyping. 
    In vine grape cultivation, flower detection suffers from challenges such as very small targets (\Cref{fig:DifficultAnnotation_a}) and partial occlusion caused by leaves (\Cref{fig:DifficultAnnotation_b}), whereas complex shapes and the difficulty of recognition attributed to dense planting affect shoot (young branches in viticulture) detection. 
    Images captured during the day can include crops of the same variety in the background, thereby making annotation more challenging (\Cref{fig:DifficultAnnotation_c}).
    Thus, subjectivity and other uncertainty factors often influence annotation, making it extremely difficult to create well-defined and consistent training data.

\begin{figure}[H]
        \begin{subcaptiongroup}
            \phantomcaption\label{fig:DomainDiversity_a}
            \phantomcaption\label{fig:DomainDiversity_b}
        \end{subcaptiongroup}
        \includegraphics[width=1.0\textwidth]{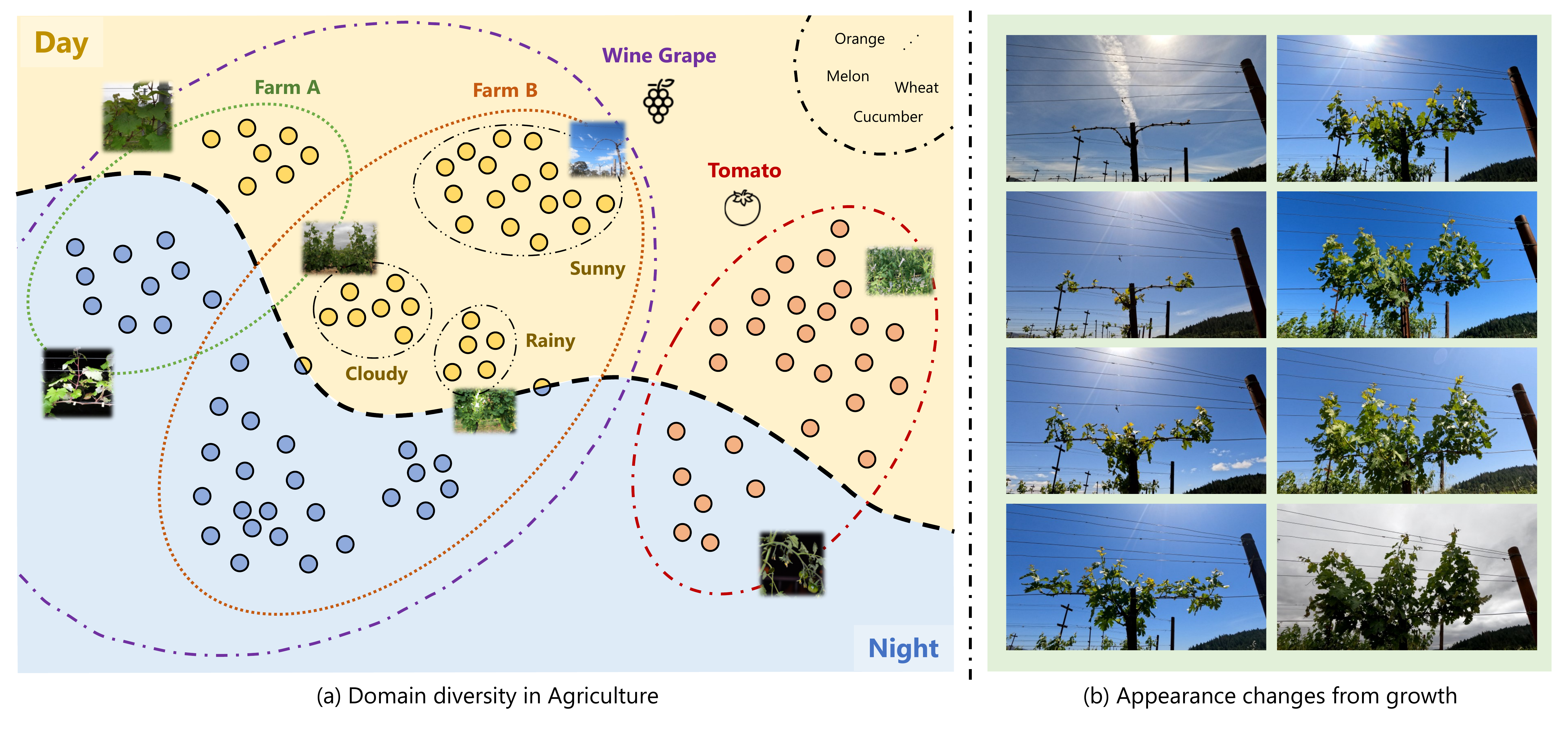}
        \caption{\textcolor{Highlight}{\textbf{Domain diversity challenges in agriculture}}\\
        (a) Domain diversity in the agricultural field\\
        (b) Changes in appearance characteristics attributed to plant growth
        }
        \label{fig:DomainDiversity}
    \end{figure}

    The diversity of the agricultural domains makes it difficult to secure object-annotated images for model training.  \Cref{fig:DomainDiversity} shows an example of domain diversity in the agricultural field (\Cref{fig:DomainDiversity_a}) and the changes in the appearance characteristics based on crop growth (\Cref{fig:DomainDiversity_b}). 

    A large-scale and diverse training dataset is required to improve the generalization performance of object- detection models. 
    This allows flexible adaptation to unknown environments in actual operational sites; however, in the agricultural field, the visual characteristics of target objects vary significantly because of climatic conditions, cultivation environments, and growth stages, thereby making it extremely difficult to collect training data that cover all these diversities. 

    In addition, although there are common plant phenotyping tasks for different crops, reusing training data is challenging because of the significant differences in appearance characteristics. 
    Consequently, models that are effective only in specific environments, conditions, or for specific crops tend to be developed, limiting their broad application in actual agricultural sites.
    
    Thus, Plant phenotyping using object detection models is a promising technology in PA; however, in actual operation, it faces several challenges in securing object-annotated images. 
    Therefore, the generalization performance of object detection models is a concern caused by overfitting, and there are limitations in improving the generalization performance and accuracy. 
    Efficiently acquiring training data and developing an operational method that allows the training data to be applied to other fields and crops is necessary to address these challenges. 

	This study proposes a new data-augmentation method (D4) that uses a text-guided diffusion model for object detection. 
    The technology for the periodic and long-term capture images of crop growth has improved with the development of automatic shooting equipment in the agricultural sector, thereby making it easier to collect large amounts of video data \citep{tokekar_sensor_2016}. 
    We propose a data-augmentation method that utilizes a pre-trained multimodal image generation model (text-guided diffusion model) with many original images extracted from video data and a small number of object-annotated images. 
    A text-guided diffusion model was used for generating new object-annotated images with background information adapted to the target domain, while retaining the annotation information necessary for object detection. 
    D4 is expected to contribute to addressing the challenges of plant phenotyping in PA and the realization of a sustainable food production system.

    The remainder of this paper is organized as follows: 
    \Cref{RelatedWork} discusses related work.
    \Cref{MaterialsAndMethods} describes D4.
    \Cref{Experiments_and_Results} evaluates D4 through experiments.
    \Cref{Disscussion} presents an analysis of the experimental results.
    \Cref{Ablation_Study} discusses the effectiveness and challenges of D4 through an ablation study.
    Finally, \Cref{Conclusion} concludes the paper.
%

\section{Related Work}\label{RelatedWork} 

    \subsection{Generative Data Augmentation for Image Recognition}
        In training deep learning models, data augmentation techniques are commonly applied when there is a small amount of training data or a need for more diversity in the dataset. 
        Data augmentation increases the diversity of the training data by adding new data, making it possible to improve the generalization ability of deep learning models and effectively and efficiently enhancing their accuracy for unknown data~\citep{wong_understanding_2016}~\citep{taylor_improving_2018}.
        In areas where it is difficult to obtain training data, such as medical image analysis~\citep{litjens_survey_2017} and agricultural image processing~\citep{consistent_2022}, data augmentation contributes to the construction of high-precision models from limited training data. 
        For image recognition tasks, data augmentation can be classified broadly into the ``data transformation approach'' and ``data synthesis approach'' ~\citep{mumuni_data_2022}.
        The data transformation approach involves data augmentation, which creates new images by transforming existing images.
        In particular, rotating or flipping images can improve the robustness of the model by enabling it to recognize objects from various positions and angles. 
        These methods are relatively simple; however, they are known to contribute significantly to improving model accuracy. 
        The application scope of the data transformation approach is limited because the generated images are restricted to variations in the original dataset. 
        Their inability to handle diverse domain changes is specified as a disadvantage.

        In contrast, the data synthesis approach involves generative data augmentation, which generates new images. 
        The accuracy of object recognition under unknown environments and conditions beyond the range of representations within a limited dataset can be improved by using image generation models such as 3DCG, GAN~\citep{GAN_2014}, and VAE~\citep{VAE_2022} for data augmentation. 
        Therefore, generative data augmentation methods of the data synthesis approach have attracted considerable attention for addressing the issue of domain diversity.

    \subsection{Generative Data Augmentation for Domain Adaptation}
        A domain shift occurs when there is a discrepancy between the distributions of the training and test data, thereby leading to a decrease in the generalization performance of the model.
        One approach to address this issue is the study of generative data augmentation techniques through data synthesis using image generation models. 
        The object recognition accuracy in unknown environments or under different conditions can be improved beyond the range of representations within a limited dataset using image generation models. 
        Several image generation models have been proposed, for example, ARIGAN proposed by Valerio et al.~\citep{valerio_giuffrida_arigan_2017}, and EasyDAM proposed by Zhang et al.~\citep{zhang_easy_2021}\citep{EasyDAM_V2}\citep{EasyDAM_V3}\citep{EasyDAM_V4}. These studies demonstrated the effectiveness of generative data augmentation using images generated in the agricultural field using image generation via GANs.
        EasyDAM constructs an image transformation model using an Across-Cycle GAN, which is an improved version of the CycleGAN~\citep{zhu_unpaired_2017} for achieving domain adaptation to different varieties. 
        Moreover, Trabucco et al. proposed DA-Fusion~\citep{trabucco_effective_2023}, which uses a diffusion model of Text2img to generate images from text, improving the generalization performance of models in classification tasks. 
        Thus, generative data augmentation methods based on the data-synthesis approach achieve domain adaptation using image-generation models, thereby significantly improving the adaptability of models to different domains. 
        Several methods achieved domain adaptation in object detection tasks using image generation models by transforming the labeled data.
        However, to the best of the authors’ knowledge, there is no generative data augmentation method for object detection tasks based on a data synthesis approach because the generated images require coordinate information indicating the location of the objects to be detected, and it is difficult to generate image-coordinate information pairs. 
        Therefore, it is necessary to explore a new data-augmentation method based on a data-synthesis approach that considers domain shifts in object detection.
        
    \subsection{IQA Metrics for Image Generation}
        Generative data augmentation using image-generation models requires high-quality images. 
        Therefore, the image quality assessment (IQA) \citep{IQA_2004} metric is used to evaluate the quality of the generated images. 
        The IQA metric is an indicator for quantitatively evaluating the quality of the generated images, and its calculation methods can be classified into three approaches: full-reference (FR), no-reference (NR), and distribution-based (DB) methods.
        The FR methods of IQA metrics evaluate the quality by directly comparing the reference and generated images.
        Examples of FR-IQA metrics include classical methods such as the peak signal-to-noise ratio (PSNR) and structural similarity index measure (SSIM) \citep{SSIM_2004}. 
        Recently, IQA metrics that use deep learning have been proposed, such as LPIPS \citep{LPIPS_2018} and DreamSim \citep{DreamSim_2023}. 
        FR-IQA metrics can be used to evaluate the quality and structural features of the generated image by comparing them with the reference image.
        The NR methods of IQA metrics evaluate the quality based on the generated image alone. 
        Examples of NR-IQA metrics include total variation (TV) and blind/referenceless image spatial quality evaluator (BRISQUE) \citep{BRISQUE_2012}. 
        The NR-IQA metrics can evaluate quality even without a reference image because they assess quality based solely on the generated image.
        Distribution-based (DB) methods of the IQA metrics evaluate the quality by comparing the distribution of image features in a dataset. 
        Examples of DB-IQA metrics include Fréchet inception distance (FID) \citep{FID_2017} and kernel inception distance (KID) \citep{KID_2021}. 
        The DB-IQA metrics can simultaneously evaluate the image quality and diversity by comparing the distribution of image features across the entire dataset. 
        The IQA metrics make it possible to quantify the quality of the generated images.  
        Therefore, we ensure the quality of the generated images during generative data augmentation by objectively evaluating and utilizing IQA metrics in this study. 
%
\section{Materials and methods}\label{MaterialsAndMethods} 

    \subsection{Dataset Overview}\label{DatasetOverview}
\begin{figure}[H]
            \begin{subcaptiongroup}
                \phantomcaption\label{fig:GrapeDataset_a}
                \phantomcaption\label{fig:GrapeDataset_b}
                \phantomcaption\label{fig:GrapeDataset_c}
            \end{subcaptiongroup}
            \includegraphics[width=1.0\textwidth]{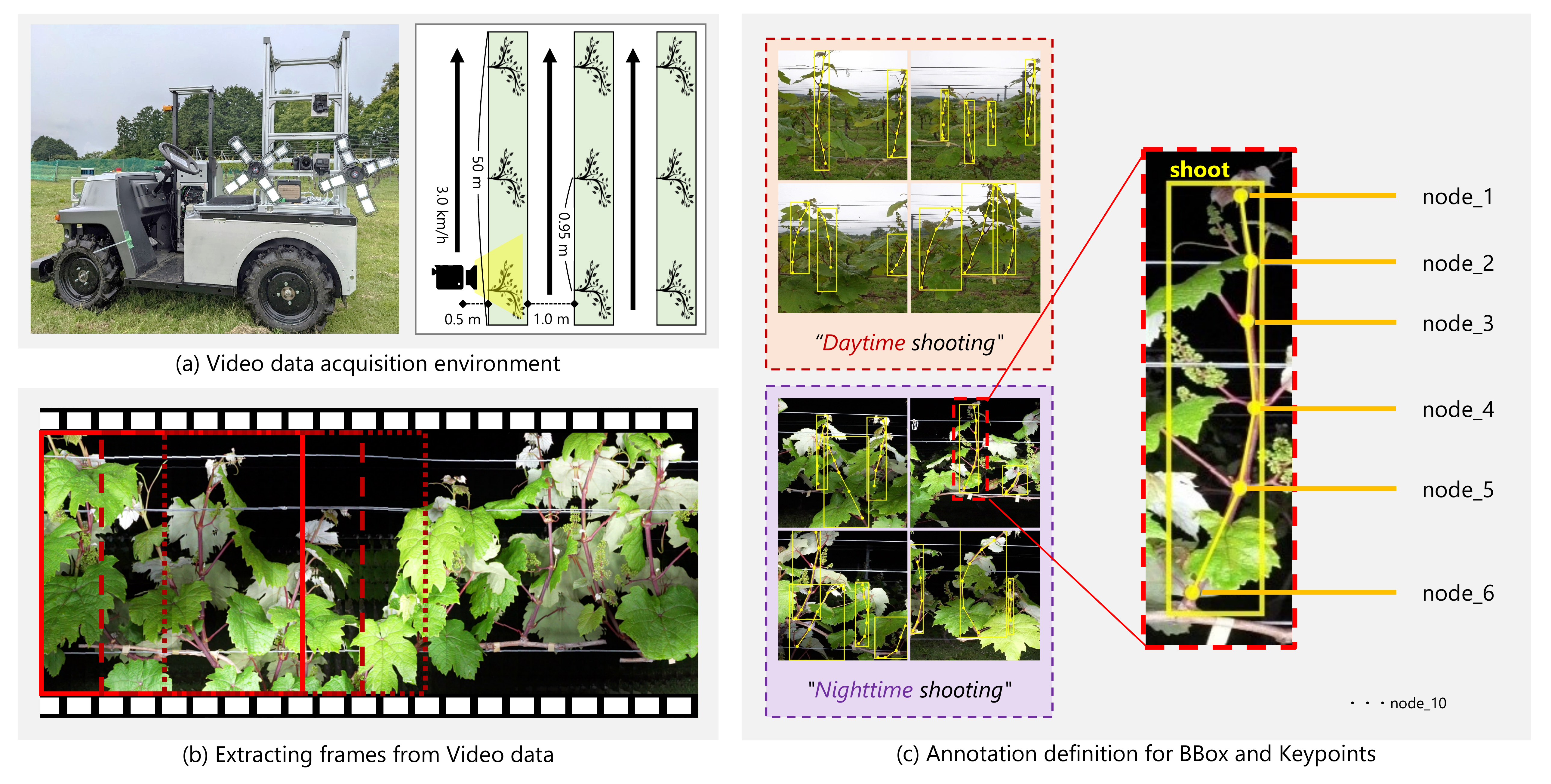}
            \caption{\textcolor{Highlight}{\textbf{Overview of the dataset used in this study}}\\
                (a) Data collection environment\\
                (b) Frame extraction from video data\\
                (c) Annotation definitions for BBox and keypoints
            }
            \label{fig:GrapeDataset}
        \end{figure}

        We prepared a custom dataset for estimating the internodal distance of shoots (young branches in viticulture) in vineyards.  
        This dataset was created to estimate the growth of grapevines at the pre-blooming BBCH stage \citep{lorenz_growth_1995} of 57--59 by capturing videos using an unmanned ground vehicle (UGV).  
        We aimed to quantify crop growth by extracting images from video frames and calculating the internodal distance through image recognition.
        An overview of the dataset is shown in \Cref{fig:GrapeDataset}. 
        The data collection was conducted using an UGV that autonomously traversed the field and captured images of grapevines from the side (\Cref{fig:GrapeDataset_a}). 
        Images were captured both during the day and at night, with artificial lighting used for nighttime photography. 
        Frames were extracted from the captured videos to create an image dataset (\Cref{fig:GrapeDataset_b}). 
        
        For object detection, annotations were defined for the BBox and keypoints (\Cref{fig:GrapeDataset_c}). 
        The BBox annotation consisted of only one class, ``Shoot.'' The Shoot class contains a keypoint label indicating ``node.'' The node labels were defined from node\_1 to node\_10 and arranged from the top to the bottom of the shoot. 
        If there were less than ten node labels, they were not annotated. As shown in \Cref{fig:GrapeDataset}, the daytime images present a challenge for annotation because of their complex background information. 
        In contrast, nighttime images captured using light sources have less background noise, which makes the annotation process easier. 
        Therefore, this study aimed to train an object detection model for daytime images using only a few annotated nighttime images. 
        We expect to develop a highly accurate detection model for daytime images by focusing on creating a small but perfect and consistent set of annotations for certain nighttime images.

    \subsection{Requirements}
        We considered generative data augmentation to simultaneously address the challenges of ``insufficient training data'' and ``diversity of domains'' for employing plant phenotyping in PA using object detection models. 
        We set the following four requirements to propose the new generative data augmentation method. 
        The D4 method aims to address the challenge of securing training data for learning object detection models and realizing plant phenotyping in PA by meeting the following requirements.
        
        \begin{itemize}
            \item Utilization of video data\\
                In modern agriculture, task automation in vast cultivation fields using an UGV has helped address labor shortages. 
                UGVs that move freely throughout the cultivation fields are gaining attention as data collection equipment for PA and used for video recording and sensor data accumulation \citep{UGV_2020}. 
                Therefore, although securing training data for image recognition in the agricultural sector is challenging, the environment for collecting cultivation data is established steadily, thereby making it easier to collect periodic video data. 
                Therefore, we consider utilizing the vast amount of video data UGV has captured and accumulated in the cultivation fields.

            \item Utilizing a small number of object-annotated datasets\\
                In the agricultural field, creating object-annotated datasets requires considerable effort; therefore, we cannot expect perfect and consistent object-annotated datasets. 
                We consider learning a high-precision detection model from a minimal number of perfectly created object-annotated datasets with minimal effort.

            \item Domain adaptation\\
                In agriculture, it is extremely difficult to create an object-annotated dataset that covers various domains. 
                Further, transferring a training dataset between different crops with the same learning task is challenging, and therefore, we consider generating object-annotated images of different domains prepared in advance without annotating the target domain.

            \item Preservation of annotation information for detection models\\
                Training detection models requires object coordinate information indicating the positions of the objects in addition to image data. 
                Therefore, exploring methods for generative data augmentation that accurately preserve the relationship between the detection targets and their coordinate information is necessary. 
                The goal is developing methods that apply not only to BBox detection but also to keypoint detection and segmentation tasks.
        \end{itemize}

    \subsection{Proposed method: D4}
        We propose a new generative data augmentation method for object recognition tasks, called ``D4,'' based on a text-guided diffusion model. 
        D4 is defined as text-guided diffusion model-based domain-adaptive data augmentation for object detection. 
        The text-guided diffusion model is a multimodal model that can transform an input image into different images under arbitrary conditions based on textual information (prompts) while preserving its structure, which is an improvement over the image-to-image model. 
        We developed a method for training data expansion and domain adaptation through a ``data synthesis'' approach in data expansion Using such an image generation model.
        
\begin{figure}[H]
            \includegraphics[width=1.0\textwidth]{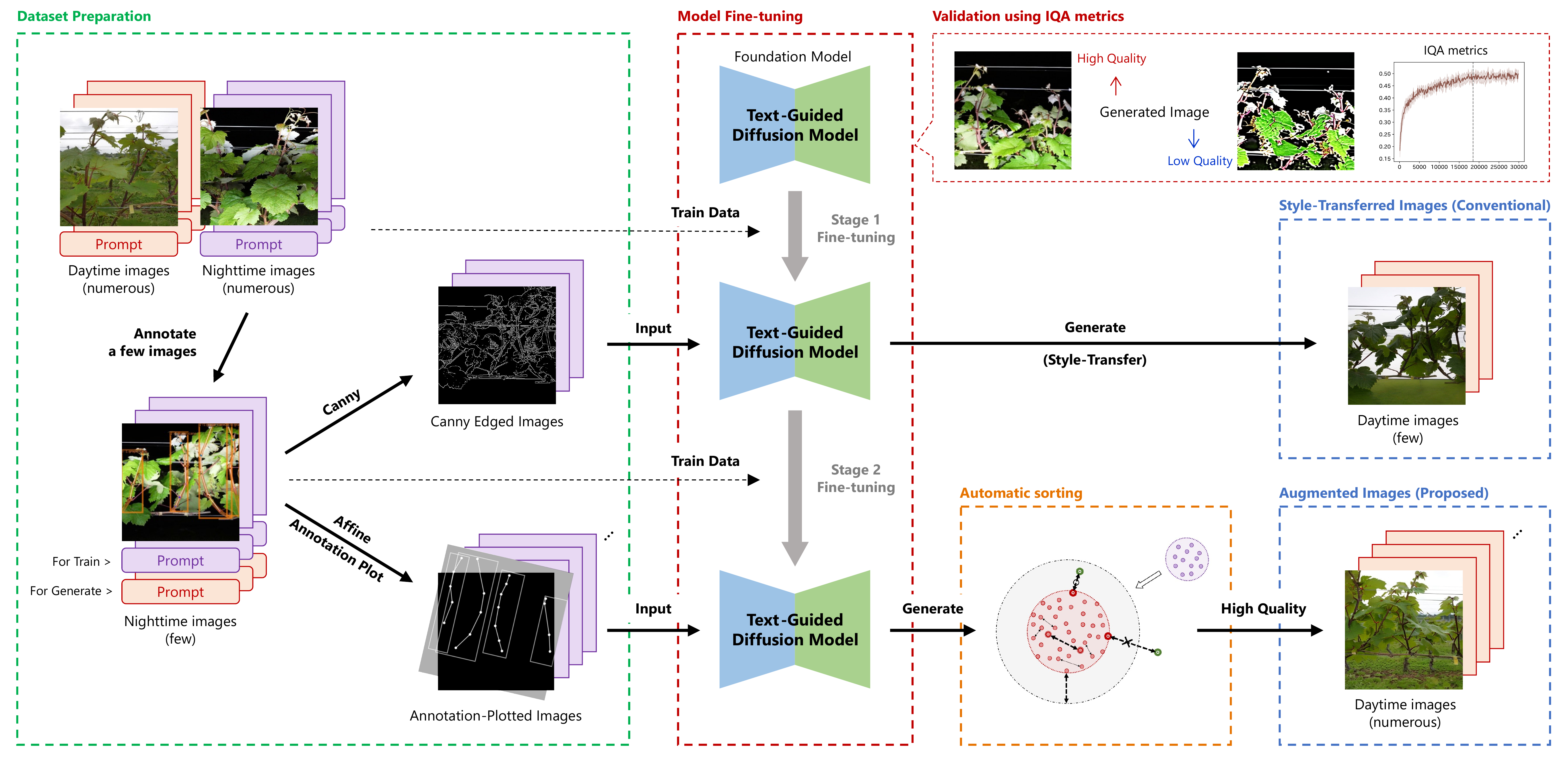}
            \caption{\textcolor{Highlight}{\textbf{Basic framework and key components of D4}}}
            \label{fig:overallStructure}
        \end{figure}

        \Cref{fig:overallStructure} presents the basic framework and key components of D4. 
        D4 uses a large dataset of real images from different domains with prompts and a small annotated dataset. 
        The prompt contains textual information indicating each image domain, and it can either be predefined fixed text or automatically generated text obtained using methods such as CLIP\citep{CLIP_2021} and GPT-4V\citep{GPT4V_2023}.
        In D4, a small annotation dataset does not need to include the target domain; further, a large number of high-quality synthetic images of the target domain can be generated based only on the annotated images of other domains. 
        This can be achieved by the two-stage fine-tuning of the text-guided diffusion model.

        In stage 1 fine-tuning, the text-guided diffusion model is trained using unlabeled data to learn broad features for the unknown dataset that is yet to be acquired by the underlying model. 
        Conventional domain adaptation using image-generation models corresponds to style transfer using a pretrained model from stage 1 \citep{StyleTransfer_2023_CVPR}\citep{StyleDiffusion_2023}.

        In stage 2 fine-tuning, an image plotted with annotation information (annotation-plotted image) is used as input to learn the local features from a small annotated dataset. \\
        In the trained model in stage 2, fine-tuning the model trained in stage 1 enables generating many high-quality images for the target domain without preparing annotated images for the target domain. \\
        In addition, the input is an annotation-lotted image, and therefore, adding geometric changes can increase variation, thereby enabling diverse image generation that cannot be achieved only using style transfer.
        Subsequently, in D4, validation based on IQA metrics is performed during both the training and generation phases of the ctext-guided diffusion model for maximizing the quality of the generated images. 
        The images generated by generative data augmentation using D4 are of high quality, equivalent to real images, and closely resemble the target domains. 
        This enables the generative data augmentation of D4 to achieve a high-accuracy improvement and generalization performance for training the object-recognition models.
        
        Subsequently, the details of pretraining the text-guided diffusion model are described in \Cref{ProposedMethod_pre_train}, details of the generative data augmentation are provided in \Cref{ProposedMethod_data_aug}, and improvement in the quality of the generated images discussed in \Cref{ProposedMethod_image_Quality}.

    \subsection{Pre-training of the text-guided diffusion model}\label{ProposedMethod_pre_train}
        \subsubsection{Overview}
\begin{figure}[H]
                \begin{subcaptiongroup}
                    \phantomcaption\label{fig:2StagePretraining_a}
                    \phantomcaption\label{fig:2StagePretraining_b}
                \end{subcaptiongroup}
                \includegraphics[width=0.7\textwidth]{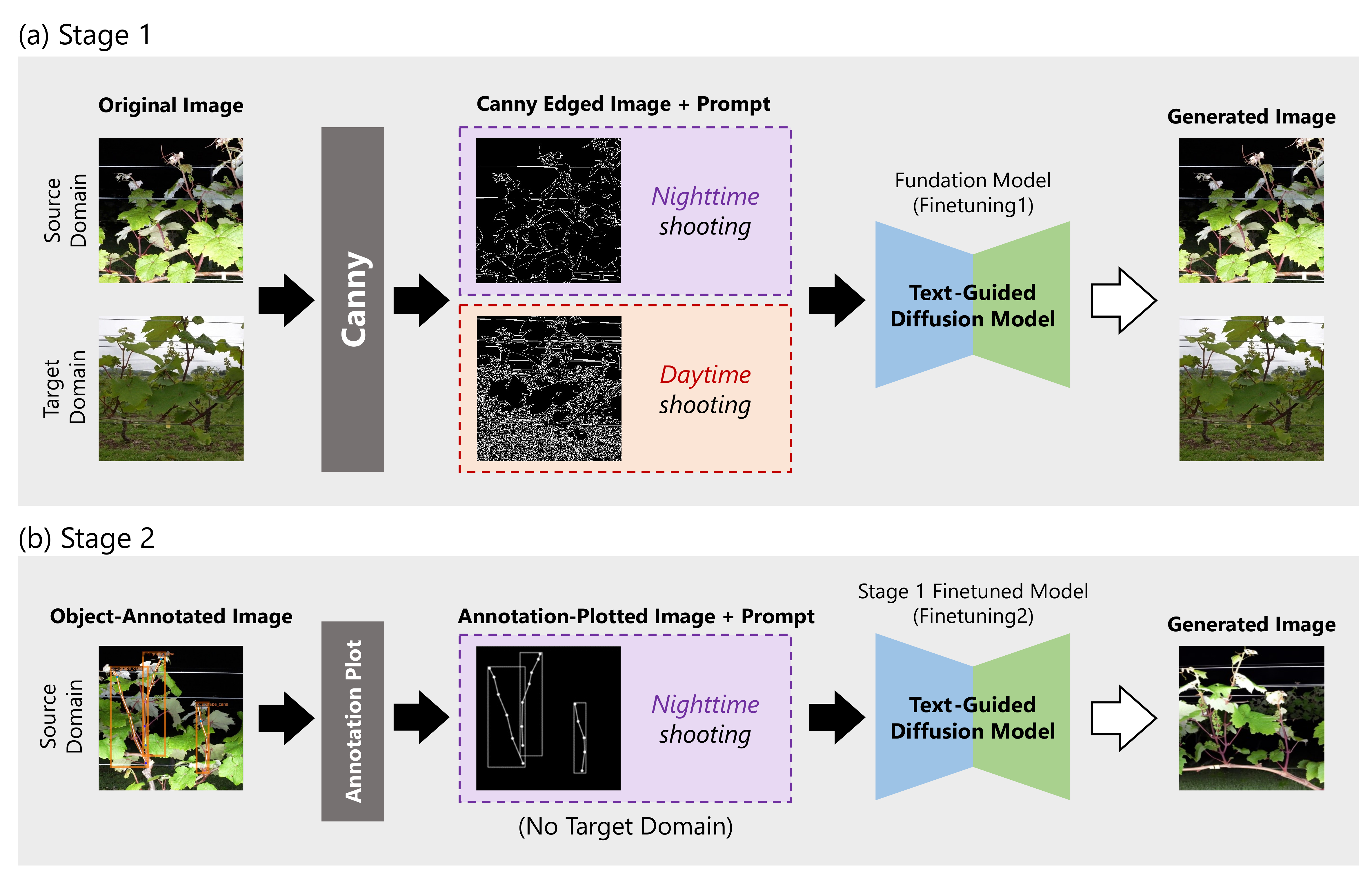}
                \caption{\textcolor{Highlight}{\textbf{Pre-training the text-guided diffusion model}}\\
                (a) Stage 1: Learning broad features in the proprietary dataset\\
                (b) Stage 2: Learning local features in the proprietary dataset
                }
                \label{fig:2StagePretraining}
            \end{figure}

            D4 is realized by pre-training the foundation model of the text-guided diffusion model with a custom dataset that includes images of the target domain for which generative data augmentation is desired.
            \Cref{fig:2StagePretraining} presents the training procedure of the text-guided diffusion model used in D4.
            The foundation model undergoes two-stage pretraining to learn both broad and local features in the proprietary dataset.
            Further, during pre-training, images generated by the text-guided diffusion model are verified based on the IQA metrics.
            The transition of the IQA metrics during the training process is monitored to select the best text-guided diffusion model that produces high-quality-generated images.

        \subsubsection{Stage 1:Learning broad features}
            The first stage of pre-training (Stage 1) involves teaching the foundation model of a general-purpose text-guided diffusion model with broad image features from a custom dataset. 
            A summary of stage 1 is shown in \Cref{fig:2StagePretraining_a}. 
            In stage 1, the model learns to generate images that closely resemble real images from inputs comprising pairs of Canny edge images and prompts. 
            For \Cref{fig:2StagePretraining_a}, the text-guided diffusion model learns the domain ``vineyard,'' which was not learned by the foundation model. 
            Consequently, the text-guided diffusion model can generate images of the target crops and cultivation fields from the custom dataset. 

            A distinctive feature of stage 1 is that it does not require annotated training data with object annotations. 
            Therefore, learning from many images extracted from video frames is possible, thereby enabling the text-guided diffusion model to acquire image features, even in domains without training data. 
            The only requirement in stage 1 is a Canny-edge image labeled with textual information (prompts) describing each region. 

            However, Canny edge images can be created automatically. 
            Further, text labeling can be as simple as setting a predetermined sentence for each domain. 
            Moreover, text labeling can be performed using CLIP\citep{CLIP_2021} or GPT-4V\citep{GPT4V_2023} to create prompts, which are significantly less expensive than object annotation.

        \subsubsection{Stage 2:Learning of local features}
            The second pre-training stage (stage 2) focuses on learning local features in the training dataset using the text-guided diffusion model trained in stage 1. 
            \Cref{fig:2StagePretraining_b} illustrates an overview of stage 2. 
            In stage 2, unlike the training in stage 1, learning is conducted for generating realistic images from inputs consisting of plain images with annotated data drawn on them (annotation-lotted images) and prompts. 
            For \Cref{fig:2StagePretraining_b}, the text-guided diffusion model learns the coordinate information of BBox and keypoint from the annotation-plotted images derived from the training dataset, thereby enabling the text-guided diffusion model to generate objects of detection based solely on coordinate information. 
            In addition, the learning conducted in stage 1 allows generation images from domains that were not present in the training dataset learned in stage 2.

    \subsection{Generative data augmentation by text-guided diffusion model}\label{ProposedMethod_data_aug}

\begin{figure}[H]
            \begin{subcaptiongroup}
                \phantomcaption\label{fig:ProposedMethod_a}
                \phantomcaption\label{fig:ProposedMethod_b}
            \end{subcaptiongroup}
            \includegraphics[width=1.0\textwidth]{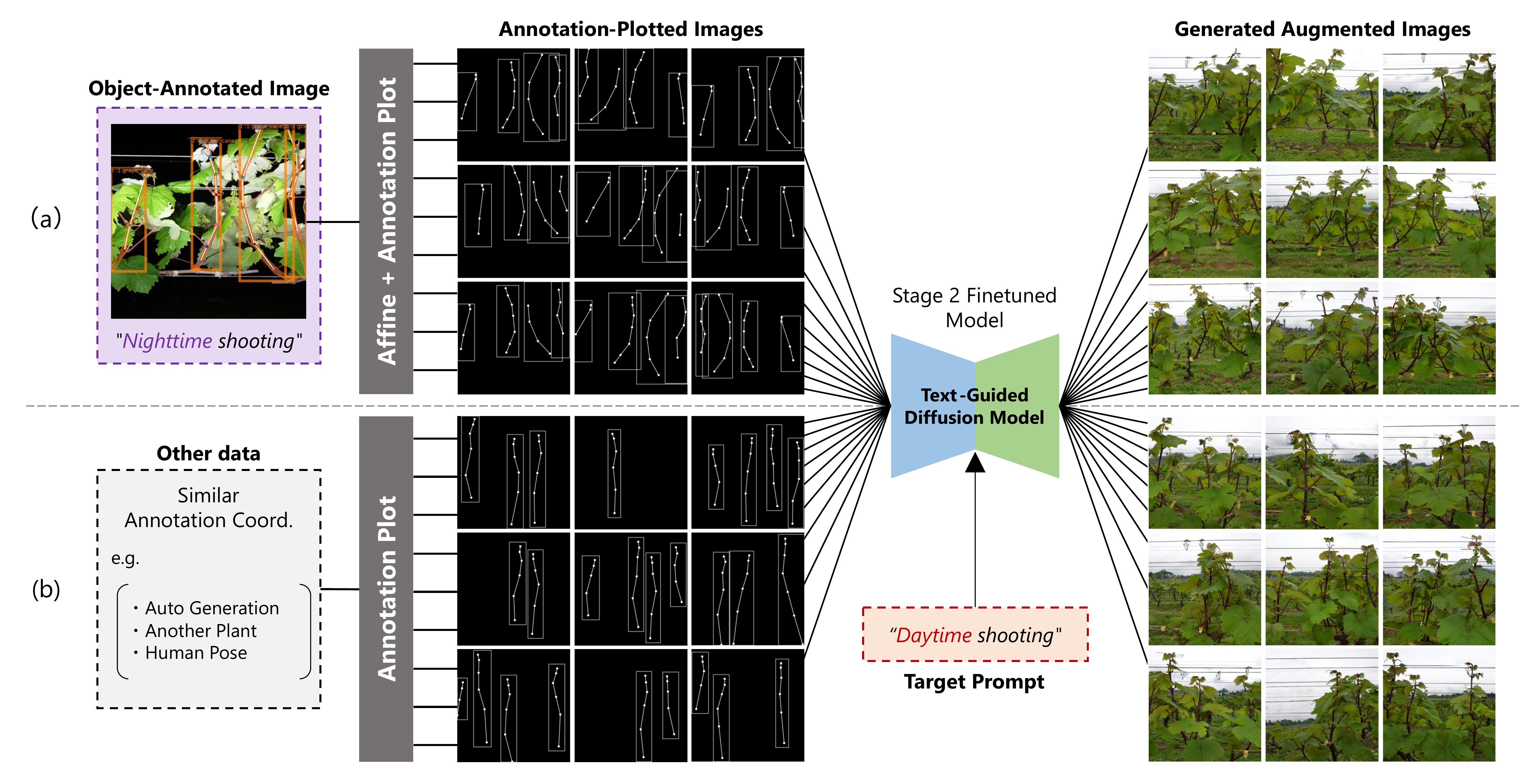}
            \caption{\textcolor{Highlight}{\textbf{Generative data augmentation via image generation using a pre-trained text-guided diffusion model}}\\
            (a) Generate nine daytime object-annotated images from one nighttime object-annotated image \\
            (b) Generate daytime object-annotated images from similar annotation coordinate information
            }
            \label{fig:ProposedMethod}
        \end{figure}

        \Cref{fig:ProposedMethod} illustrates generative data augmentation via image generation using a pre-trained text-guided diffusion model. 
        Further, pre-training the text-guided diffusion model makes it possible to shift to any domain from an annotation-plotted image based on the prompt. 
        In the figure, an annotation-plotted image is created from nighttime image training data, and daytime images are generated through prompt ``Daytime shooting.'' Thus, employing D4 makes it possible to  expand the dataset significantly under various scenarios and conditions without relying solely on limited training data or data under specific conditions. 
        Indeed, \Cref{fig:ProposedMethod_a} shows the results of generating nine daytime images from a single nighttime image using D4.

        A significant feature of this technique is that an annotation-plotted image can be created, and images of any domain can be generated from a pre-trained text-guided diffusion model. 
        Indeed, coordinating the information that matches the task of the object to be detected is necessary. 
        However, this coordinate information can be generated automatically or created from the teaching data of different crops. 
        In addition, although limited to this research task, it is possible to generate images of any domain using human skeletal information.  
        As images from entirely different domains can be used to generate images of any domain, various teaching datasets can be repurposed, thereby realizing domain adaptation. 
        \Cref{fig:ProposedMethod_b} shows the results of generating nine daytime images using nine types of coordinate information generated by a simple algorithm through D4.
        
        In addition, images generated by the image generation model were as close as possible to the real images. 
        Thus, it is possible to use them in conjunction with generative data augmentation methods of the ``data transformation'' approach, such as rotation and flipping. 
        Further improvements in the generalization performance and robustness of the object detection model can be achieved by combining them with the data transformation approach.

    \subsection{Automatic selection mechanism of generated images}\label{ProposedMethod_image_Quality}
        In D4, the quality of images generated by the text-guided diffusion model is paramount. 
        Therefore, a mechanism for automatically selecting the generated images was developed, and this automatic selection mechanism filters only high-quality images for use in generative data augmentation.

        A challenge arises when selecting the generated images. 
        This challenge lies in the discrepancy between the domains of the original and generated images. 
        For example, when converting a dataset of nighttime images into daytime images, comparing the quality of the images directly between different domains is problematic.
        DreamSim\citep{DreamSim_2023} is utilized to compare the distance in the feature space of the target domain dataset with the generated images to circumvent this problem, thereby facilitating the selection process through a quality comparison of the generated images. 
        DreamSim, leveraging embedding representations created by CLIP\citep{CLIP_2021} and DINO\citep{DINO_2021}, is a FR-IQA metric that can capture not only foreground objects and semantic content with an emphasis, but also texture and structural features.
        Further, using embedded representations simplifies the identification of similar images, and therefore,  DreamSim was employed as an automatic selection mechanism for the generated images because it can easily extract the most similar images from the target dataset.

\begin{figure}[H]
            \includegraphics[width=0.7\textwidth]{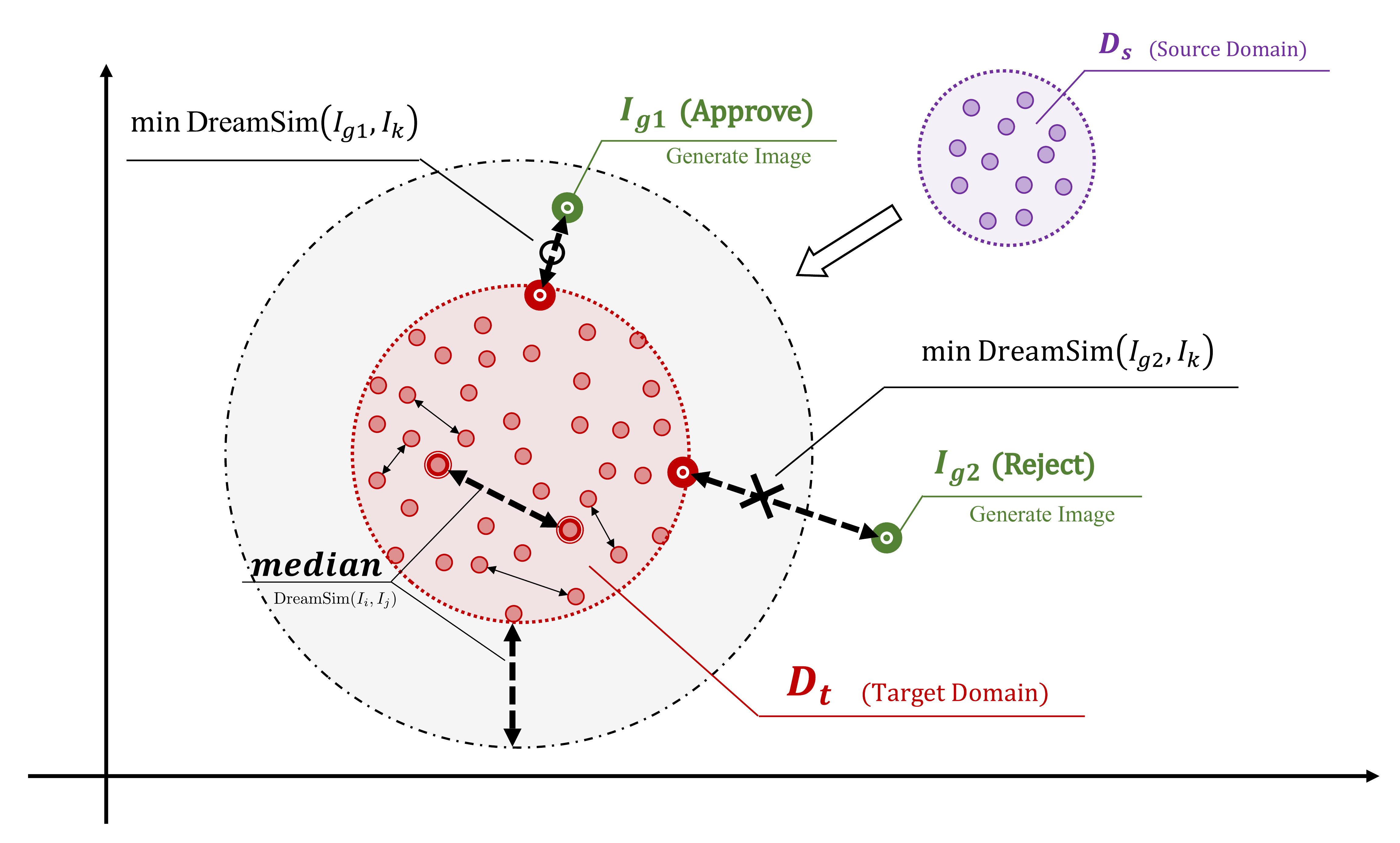}
            \caption{\textcolor{Highlight}{\textbf{Automatic selection mechanism for generated images using DreamSim}}}
            \label{fig:DreamSimAutoSelection}
        \end{figure}

        \Cref{fig:DreamSimAutoSelection} shows the generated image selection mechanism using DreamSim. 
        Conditions for determining the generated images are given in \Cref{calc:DreamSimAutoSelection}. 
        Using this selection mechanism, the DreamSim score between any pair of images within the dataset is calculated, and the median is determined. 
        The generated image is considered to be of high quality if the DreamSim score between a generated image and most similar image in the dataset is higher than the median. 
        Only images deemed high-quality by this selection mechanism were used as augmented data, thereby ensuring a certain quality in the generated images augmented in D4.

        \begin{equation}
            \label{calc:DreamSimAutoSelection}
            Quality(D_t, I_g) = \begin{cases} 1 & \text{if } \text{median}\{\text{DreamSim}(I_i,I_j) | i, j \in D_t, i \neq j\} > \min\{\text{DreamSim}(I_g,I_k) | k \in D_t\} \\0 & \text{otherwise} \end{cases} 
        \end{equation}

%
\section{Experiments and Results}\label{Experiments_and_Results} 

    \subsection{Overview of experiment}
        \subsubsection{Purpose}
            In our experiments, we tested the effectiveness of domain adaptation in object detection tasks using the proposed D4. 
            Our goal was quantitatively evaluating the effectiveness of D4 in both the BBox and keypoint detection models. 
            In these experiments, generative data augmentation was performed solely on a small dataset of nighttime images, and its effectiveness was evaluated against a daytime image test dataset. 
            Through this experiment, we provided quantitative results demonstrating the effect of limited training generative data augmentation and domain adaptation in D4.

        \subsubsection{Base datasets}
            This experiment uses a dataset for estimating the internodal distances in grapevine cultivation, as shown in \Cref{DatasetOverview}. 
            The dataset for training the detection model comprises 100 nighttime images, whereas the test dataset comprises 50 daytime images. 
            We aimed to improve the accuracy of the detection model and quantitatively evaluate the effectiveness of D4 based on the accuracy of the model by implementing generative data augmentation and domain adaptation using D4.

            An original image dataset, which did not require annotations, was prepared to implement D4. 
            This dataset includes daytime and night-time image domains, which were adjusted to contain 10,352 images each, totaling 20,704 images. 
            All images were resized to a uniform resolution of 512 pixels (height and width).
        
        \subsubsection{Text-guided diffusion model}
            ControlNet\citep{ControlNet_2023}, which is a multimodal image generation model based on a diffusion model founded on StableDiffusion\citep{StableDiffusion_2022}, was employed as the text-guided diffusion model in the experiments. 
            This model enhances the controllability and flexibility of image generation significantly by inputting image and text information simultaneously. 
            In the experiments, the image was generated from coordinate information during the daytime by performing two-step pretraining with ControlNet; further, the effectiveness of D4 in the object detection models was evaluated.

        \subsubsection{Prompts}\label{Experiments_Prompts}
\begin{figure}[H]
                \includegraphics[width=1.0\textwidth]{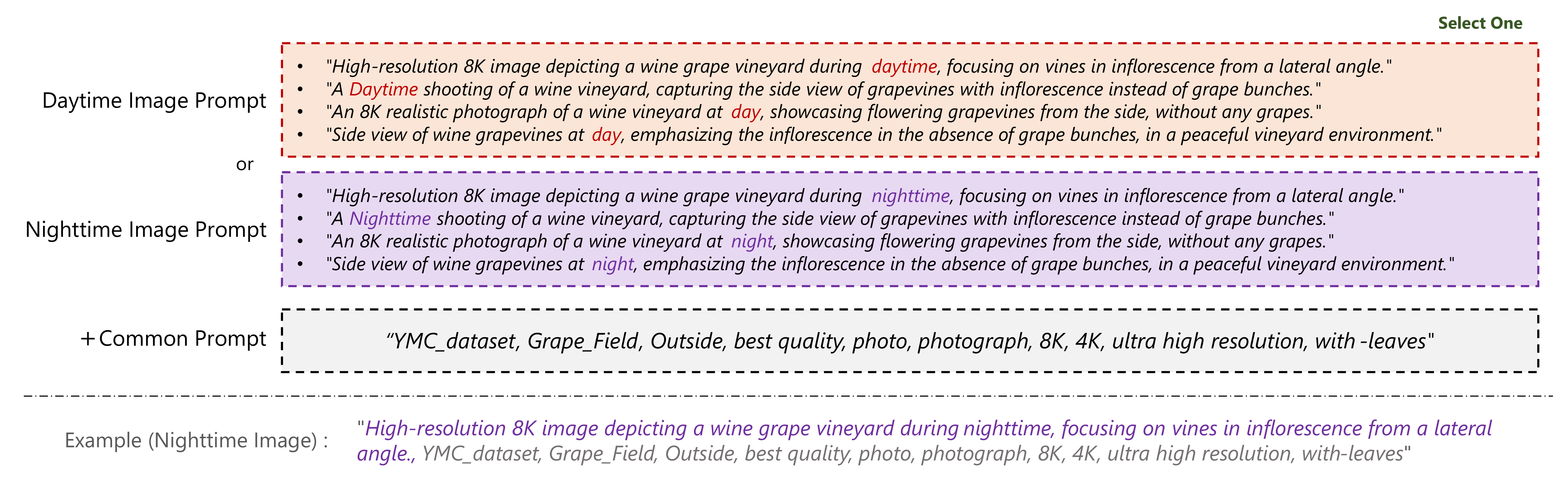}
                \caption{\textcolor{Highlight}{\textbf{Prompt used in this experiment}}}
                \label{fig:Prompts_w_PE}
            \end{figure}

            The prompts used in the experiment combined a common prompt for all images with two different texts for the daytime and nighttime images. 
            Figure \ref{fig:Prompts_w_PE} shows prompts used in the experiments. 
            We based the prompts on the outputs of Img2Text models such as CLIP and GPT-4V. 
            This approach is grounded in prompt engineering, where using meaningful sentences rather than merely listing words can enhance the quality of the generated images.

    \subsection{Pre-Training of ControlNet}
        \subsubsection{Stage 1 Pre-Training of ControlNet}
            In stage 1, we fine tune the publicly available ControlNet foundation model and train it to generate realistic images from the inputs of Canny edge images linked with prompts. 
            We prepared four parameters with varying edge densities to create the Canny edge images, thereby allowing ControlNet to demonstrate high generalization capabilities across images with different edge densities, ultimately enabling the generation of high-quality images. 
            In addition, the training dataset was augmented by a factor of two via horizontal flipping data augmentation.  
            Therefore, we prepared 165,632 pairs of inputs and the corresponding images associated with prompts to train ControlNet.  
            The training of ControlNet was conducted with a batch size of 8 and a learning rate of 3e-4.  
            The training continued for up to 30,000 steps, with validation based on the IQA metric conducted every 100 steps. 
            According to the validation, the model at step 18,500, which showed the best value of LPIPS\citep{LPIPS_2018}, was used for training in stage 2.  
            For further details, see the Appendix (\Cref{ControlNet_Stage1_Appendix}).

        \subsubsection{Stage 2 Pre-Training of ControlNet}
            In stage 2, the ControlNet model pre-trained in stage 1 was fine tuned, and the prompts associated with the annotation-plotted images created from object-annotated images were input to train the model for generating realistic images. 
            As the training dataset, a set of 100 input-answer pairs was prepared by applying horizontal flipping data augmentation to 50 object-annotated nighttime images, thereby doubling the dataset. 
            These pairs were used to train the ControlNet model with the associated prompts.
            
            The ControlNet model was trained with a batch size of one and a learning rate of 5e-5. 
            Setting the batch size and learning rate to values lower than those used for stage 1 training helped addressed the early convergence of training attributed to the extremely small amount of data in stage 2.
            Model validation was performed by generating images from the remaining 50 object-annotated nighttime images that were not used in the training dataset and performing validation based on IQA metrics every 50 steps. 
            The model with 1200 steps, which showed the best value in LPIPS, was selected as the image generation model for D4.
            The details are provided in the Appendix.  (\Cref{ControlNet_Stage2_Appendix})
            
\begin{figure}[H]
                \includegraphics[width=1.0\textwidth]{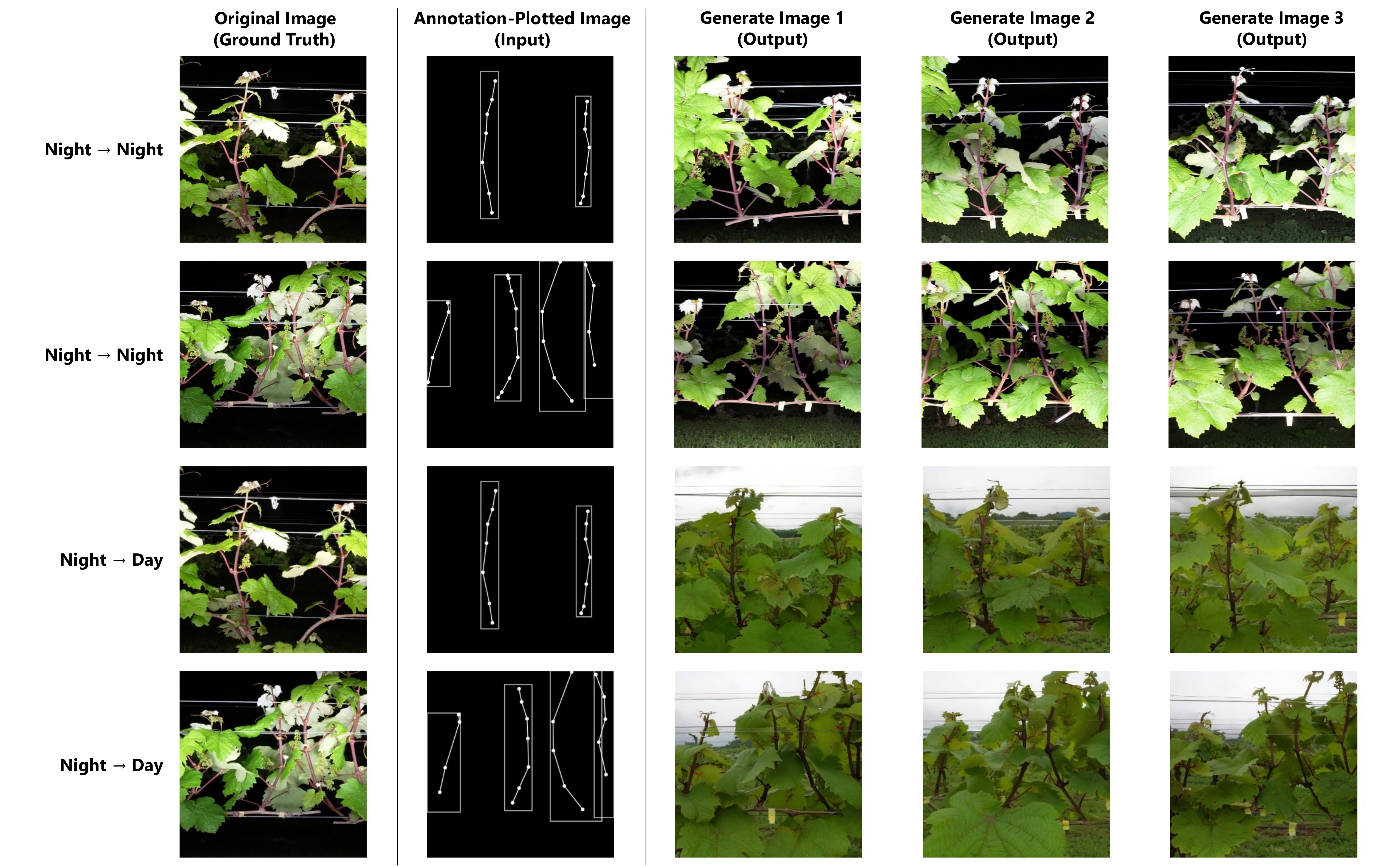}
                \caption{\textcolor{Highlight}{\textbf{Results of generating ControlNet models with stage 2 Pre-Training}}}
                \label{fig:PreTrainResult_Stage2}
            \end{figure}

            In \Cref{fig:PreTrainResult_Stage2}, we present the results generated by the ControlNet model after two stages of pretraining.  
            The input was an annotation-plotted image obtained from an object-annotated nighttime image. 
            The target objects (shot and node) in the annotated parts of the nighttime images were correctly generated when generating images using the same prompts as the original.  
            Furthermore, images were generated by inputting prompts for different daytime images. Consequently, we correctly generated daytime images and verified the matching annotations for the detected objects. 
            Therefore, the domain of daytime images was acquired through pre-training in stage 1, enabling generation without including daytime images in the training data in stage 2.  
            This suggests that the two-stage pre-training process enables domain adaptation from nighttime to daytime images through prompts.
        
    \subsection{Generative Data Augmentation by D4} 
        A dataset was created for use in the experiments via D4. 
        The dataset used for the training the detection model was augmented, and the experiments were conducted to test D4's effectiveness based on the accuracy of the detection model. 
        A total of 100 object-annotated images captured at night were prepared as the initial dataset
        This dataset was divided into 50 images for training (Normal training dataset(a)) and validation (Normal validation dataset) purposes.
        Six training datasets were prepared for comparison.
        
        First, we prepared a dataset that was style-transferred using ControlNet trained in stage 1 (Transferred Train Dataset (b)). 
        This dataset is a Canny edge image created from a real image and style transferred into a daytime image while preserving the composition using the model learned in stage 1. 
        Subsequently, five training datasets were prepared by generating images through D4 and by adding them to the ``transferred training dataset(b)." 
        These training datasets were prepared with 50, 100, 250, 500, and 1000 images augmented through D4, and they were denoted as datasets c, d, e, f, and g, respectively. 
        Performances of the models trained using each dataset was compared to evaluate the effects of D4.
        
        The normal validation dataset was used to learn the normal training dataset (a). 
        A transferred validation dataset, which ControlNet converted and learned in stage 1, was used to learn with the other datasets.

    \subsection{Experiments with BBox Detection Models}
        \subsubsection{Overview}
            In the object-detection tasks, we validated the effectiveness of D4 for the BBox detection models that estimate the rectangular information of the objects to be detected. 
            In our experiments, we evaluated three BBox detection models: YOLOv8, which specializes in high-precision, real-time BBox detection; EfficientDet, a CNN-based model capable of high-precision detection \citep{EfficientDet_2020}; and DETR, which adopts a transformer-based architecture \citep{DETR_2021}. 
            We confirmed the effectiveness of D4 by testing models with different architectural designs. 
            Training was conducted with YOLOv8 for 500 epochs and the other models for 300 epochs to evaluate and compare the model that showed the highest accuracy on the validation dataset at each epoch.
            
            For YOLOv8 training, a default-implemented data augmentation pipeline was applied. The existing data augmentation pipeline was applied, which included  distortion, scale, color space, crop,  
            flip, rotate, random erases \citep{RandomErasing_2020},  cutout \citep{Cutout_2017},  hideandseek \citep{HideandSeek_2017},  GridMask ~ \citep{GridMask_2020}, Mixup \citep{mixup_2018}, CutMix \citep{cutmix_2019}, and mosaics \citep{yolov4_2020}.  
            Thirteen types of existing data augmentation techniques were applied automatically during training. 
            EfficientDet/DETR was trained without applying existing data augmentation methods.

        \subsubsection{Results}
            In our experiments, we applied three types of BBox detection models across seven different datasets and conducted training under 21 different conditions. 
            For each condition, we ensured the reliability of the results by conducting training at least five times with various seed values and by calculating the average and standard deviation of the evaluation metrics.
            We used the mean average precision (mAP) based on IoU thresholds as an evaluation metric to assess the effectiveness of D4 quantitatively. 
            The experimental results of the BBox detection models are listed in Table \Cref{table:BBoxResults}. 
            Three main trends are evident from the results.

            First, the effectiveness of D4 is discussed. 
            Datasets c, d, e, f, and g augmented using D4 demonstrated an improvement in the detection accuracy across all BBox detection models compared with that of the the regular training dataset (a). 
            This implies the success of domain adaptation from nighttime to daytime images through D4.

            Second, the improvement in accuracy is associated with an increase in the number of data augmentation images because of D4. 
            In every BBox detection model, with an increase in the number of images in the dataset, the highest detection accuracy was observed for the dataset of 1050 images (g) in terms of mAP50. 

            In the YOLOv8 results, the 1050 image dataset (g) achieved an improvement in accuracy of 13.2\% in mAP and 27.69\% in mAP50 compared to the regular training dataset (a). 
            In contrast, for the mAP values of EfficientDet, no improvement in accuracy was observed with an increase in the number of augmented images. 
            In all BBox detection models, no significant improvement in accuracy was observed in the datasets with more than 300 images (e, f, g).  
            This suggests that the effect of D4 may saturate beyond a certain number of augmented images. 
            Therefore, selecting high-quality images and enhancing their diversity are considered more important than indiscriminately adding data.

            Third, improvement in accuracy when the existing data augmentation methods of the data transformation approach are applied simultaneously. 
            A difference in the range of accuracy improvement was observed between YOLOv8, where the existing data augmentation methods of the data transformation approach were applied simultaneously, and EfficientDet and DETR were not applied. 
            When comparing the difference in the mAP50 accuracy between the regular training dataset (a) and dataset comprising 1050 images (g), EfficientDet achieved a 22.54\% improvement in accuracy, DETR achieved a 15.99\% improvement, and YOLOv8 achieved a 27.69\% improvement. 
            Therefore, the combination of D4 and the existing data augmentation methods of the data transformation approach suggests the possibility of further accuracy improvement.

            \begin{table}[H]
                \caption{\textcolor{Highlight}{\textbf{Experimental results in the BBox detection models}}}
                \includegraphics[draft=false,width=1.0\textwidth]{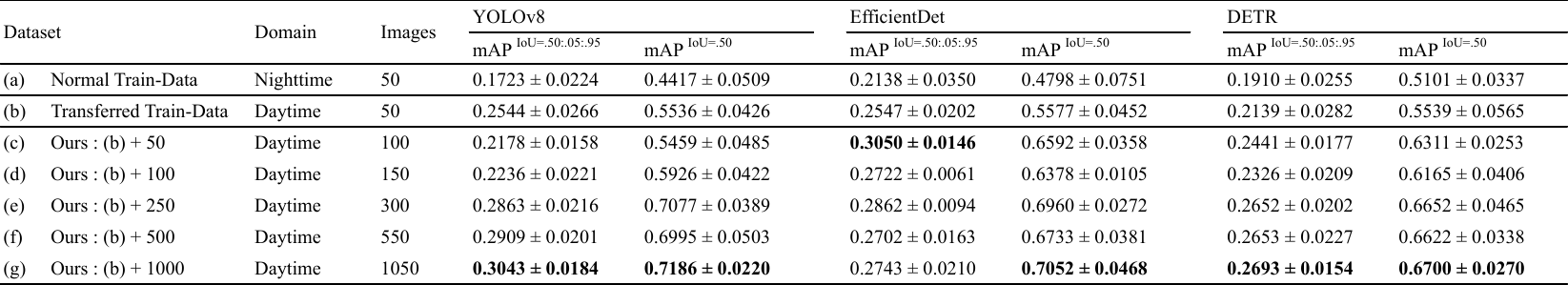}
                \label{table:BBoxResults}
            \end{table}

    \subsection{Experiments with Keypoint Detection Models} 
        \subsubsection{Overview}
            In the field of object detection tasks, we validated the effectiveness of D4 for keypoint detection models that estimate the coordinate information of the points of the detected objects. 
            In our experiments, we evaluated three types of keypoint detection models based on a top-down approach. 
            The top-down approach was adopted because the nodes of the plants have similar features, making detection difficult using the bottom-up approach. 
            The models compared were YOLOv8\_Pose, which is CNN-based and specializes in real-time detection; HR-Net\citep{HRNet_2019}, which is CNN-based and capable of high-accuracy detection; and ViT-Pose\citep{ViTPose_2022}, which adopts a Transformer-based architecture. 
            We verified the effectiveness of D4 by testing models with different architectural designs. 
            Training was conducted for 500 epochs with YOLOv8\_Pose and 300 epochs with the other models to evaluate and compare the best model that showed the highest accuracy in each epoch against the validation dataset.

            The implemented default data augmentation pipeline was applied to  train  YOLOv8\_Pose. 
            In contrast, for HR-Net and ViT-Pose, training was conducted without applying the existing data augmentation methods.

        \subsubsection{Results}
            Three types of keypoint detection models were applied to the seven different datasets. 
            The training was conducted under 21 different conditions using these combinations. 
            For each condition, we ensured the reliability of the results by conducting training at least five times with various seed values and calculating the average and standard deviation of the evaluation metrics.

            The evaluation metric utilizes the mean average precision (mAP) based on the object keypoint similarity (OKS) threshold to assess the efficacy of D4 quantitatively. 
            The formula for calculating OKS is given in \Cref{calc:OKS}. 
            In this formula, $N$ represents the number of keypoint classes, and $d_{i}$ indicates the Euclidean distance between the predicted and the actual keypoints. 
            The scale of the object is denoted by $s$, and $k_{i}$ represents a unique constant that indicates the significance of each keypoint class. 
            Furthermore, $\delta(v_{i}>0)$ represents an indicator function that equals 1 if keypoint $i$ is visible and 0 otherwise.

            Based on \Cref{calc:OKS}, the calculation of OKS requires a unique constant $k_{i}$ for each keypoint class; however, in this study, the constant $k_{i}$ was uniformly set to 0.1 for the computation. 
            Therefore, only the resulting OKS may be accurate. 
            Consequently, the keypoint detection model was evaluated based on the relative evaluation values for each dataset.

            \begin{equation}
                \label{calc:OKS}
                OKS = \frac{\sum_{i}^{N} exp(-d_{i}^{2}/2s^{2}k_{i}^{2})\delta(v_{i}>0)}{\sum_{i}^{N}\delta(v_{i}>0)}
            \end{equation}

            The experimental results for the keypoint detection models are presented in \Cref{table:KeypointsResults}.
            Three main trends are evident from the results.

            First, the effectiveness of the D4.   
            Similar to the experimental results of the BBox detection model, datasets c, d, e, f, and g augmented using D4 showed an improvement in detection accuracy across all keypoint detection models. 
            Therefore, the effectiveness of D4 is suggested for both the BBox and keypoint detection models.

            Second, the improvement in accuracy associated with an increase in the number of data augmentation images owing to D4.   
            Similar to the experimental results of the BBox detection model, we observed an improvement in the accuracy of the object detection model with an increase in the number of images in the dataset. 
            However, for HR-Net, we confirmed that the evaluation scores of mAP and mAP50 were reversed between datasets with a total of 550 (f) and 1050 images (g). 
            This result suggests that, similar to the experimental outcomes of the BBox detection model, there is a limit to the number of images that can be augmented using D4. 
            Thus, it is necessary to consider the quality of the generated images.

            Third, the results for the ViT-Pose.   
            In the evaluation of ViT-Pose, a significant improvement in accuracy was observed for a dataset with 1050 images (g) based on the mAP50 metric. 
            However, this was presumed to be caused by the inherent characteristics of the ViT model because ViT models exhibit significantly lower accuracy with extremely small amounts of data \citep{Why_ViT_2023}. 
            Therefore, D4, by increasing the amount of data, could have contributed to the performance improvement of the ViT model.

            \begin{table}[H]
                \caption{\textcolor{Highlight}{\textbf{Experimental results in keypoint detection models}}}
                \includegraphics[draft=false,width=1.0\textwidth]{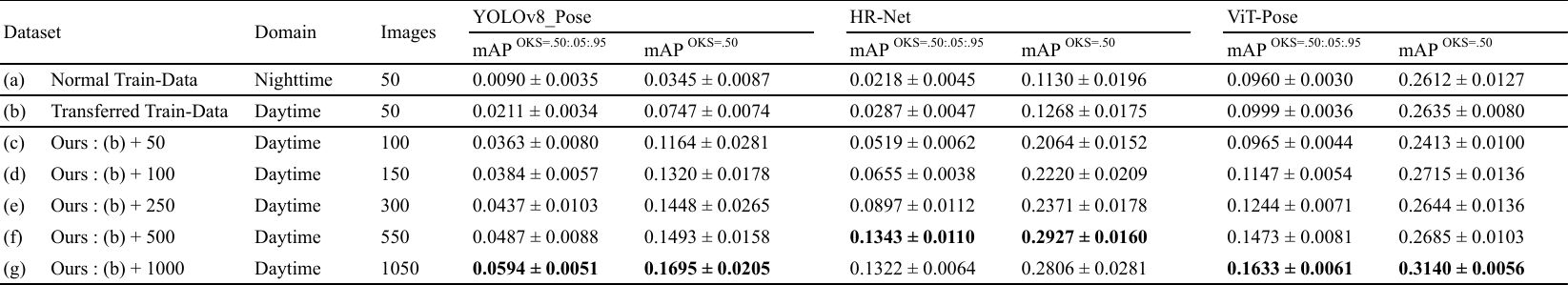}
                \label{table:KeypointsResults}
            \end{table}

%
\section{Discussion}\label{Disscussion}
    \subsection{Qualitative evaluation of generated images} 
\begin{figure}[H]
            \begin{subcaptiongroup}
            \phantomcaption\label{fig:BadGeneratedImages_a}
            \phantomcaption\label{fig:BadGeneratedImages_b}
            \phantomcaption\label{fig:BadGeneratedImages_c}
            \end{subcaptiongroup}
            \includegraphics[width=0.7\textwidth]{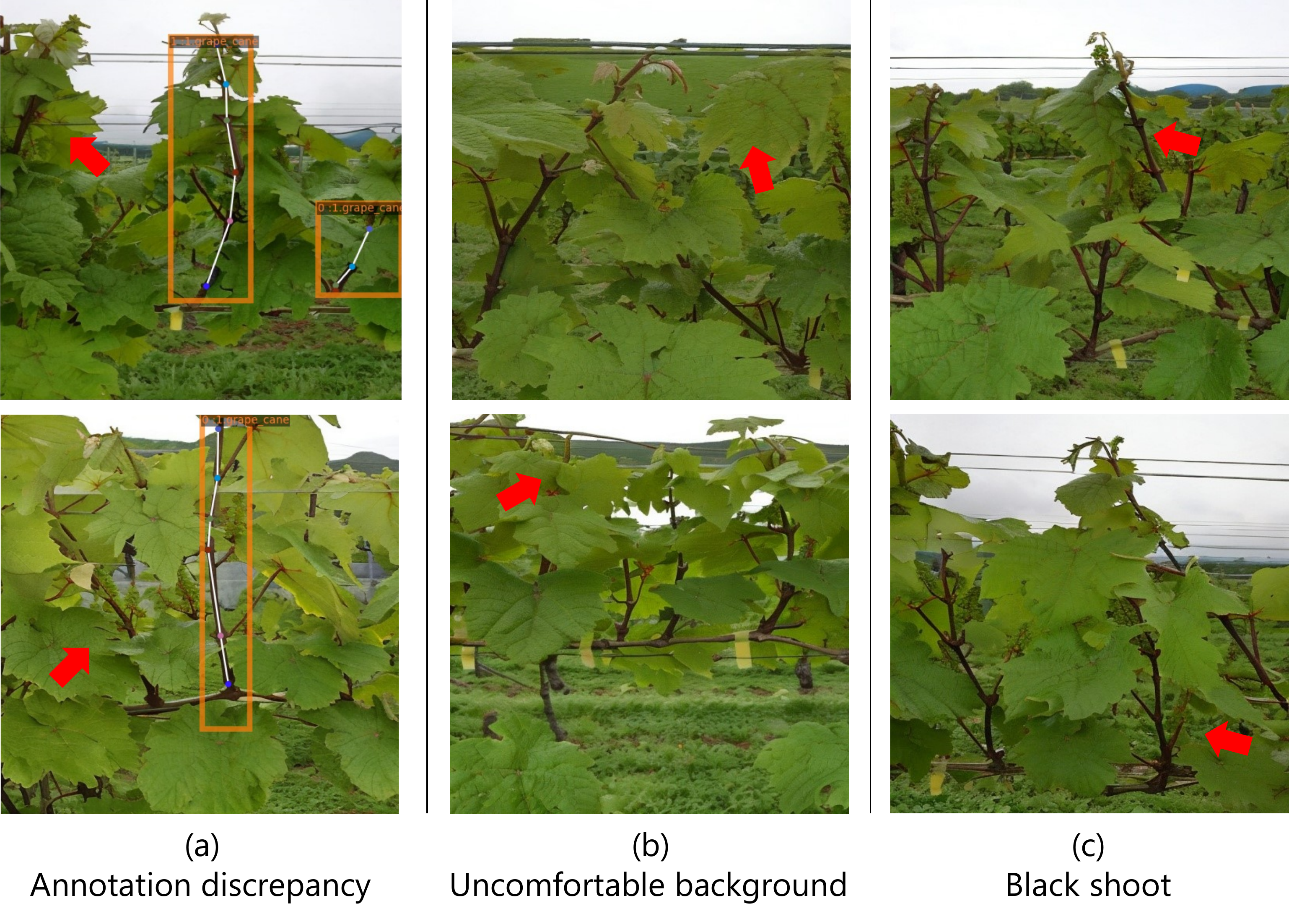}
            \caption{\textcolor{Highlight}{\textbf{Examples of generated images qualitatively assessed as poor quality}}\\
            (a) Annotation discrepancy\\
            (b) Uncomfortable background\\
            (c) Black shoot
            }
            \label{fig:BadGeneratedImages}
        \end{figure}

        The validity of images generated using D4 was assessed qualitatively based on human perceptual judgment. 
        Figure \ref{fig:BadGeneratedImages} presents examples of generated images that deemed to have poor quality in the qualitative evaluation. 
        In these poor-quality images, there are instances where the detected object and  annotation information do not match (\Cref{fig:BadGeneratedImages_a}). 
        In addition, we identified the generated images with unnatural backgrounds (\Cref{fig:BadGeneratedImages_b}) and images in which the branches were excessively dark (\Cref{fig:BadGeneratedImages_c}), clearly did not resemble real images.
        These qualitative evaluation results suggest that generated images with mismatches between the detected object and annotation information, as well as low-quality images that do not resemble real images, may function as noise in the training data of the detection model.

    \subsection{Quantitative evaluation of generated images} 
        The feature distribution of the generated images was quantitatively analyzed, uncovering discrepancies between the feature distributions of the real and generated images that were not apparent in the qualitative evaluations. 
        The visualization through dimensionality reduction and quantitative analysis using the DB-IQA metric were conducted to determine the feature distribution of the generated image dataset. 
        The datasets used for the quantitative analysis consisted of the following five types:

        \begin{itemize}
            \item G1: Test. Original image dataset (Daytime)
            \item G2: Original image dataset (Nighttime)
            \item G3: Rejected generated image dataset (Daytime)
            \item G4: Accepted generated image dataset (Daytime)
            \item G5: Original image dataset (Daytime)
        \end{itemize}

        Each dataset was randomly sampled to obtain 50 images. 
        For the visualization through dimensionality reduction, principal component analysis (PCA)\citep{PCA_1901}, t-SNE\citep{tSNE_2008}, and UMAP\citep{UMAP_2020} were employed based on the features of inception-V3\citep{szegedy_rethinking_2016}.
        For a quantitative evaluation using IQA metrics, the similarity between the test daytime real image dataset (G1) and other datasets was calculated using FID\citep{FID_2017} and KID\citep{KID_2021}.

\begin{figure}[H]
            \includegraphics[width=1.0\textwidth]{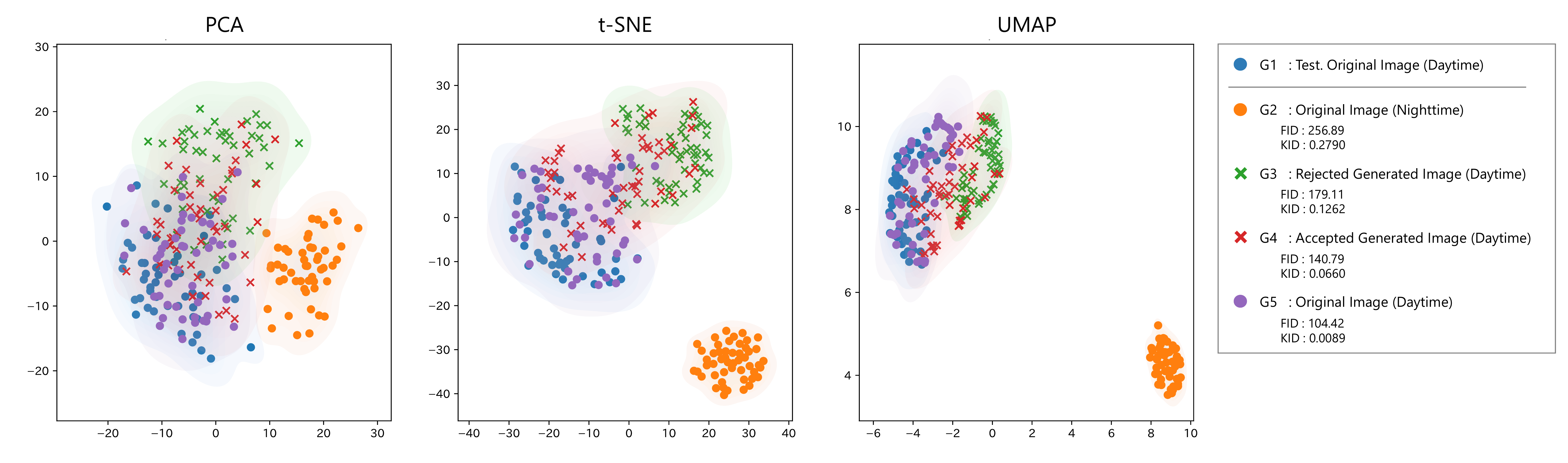}
            \caption{\textcolor{Highlight}{\textbf{Visualization of image feature distributions by dimensionality reduction}}}
            \label{fig:PlotGenImageFeatures}
        \end{figure}

        The results of the quantitative evaluation using IQA metrics are presented in \Cref{table:GenImageIQAmetrics}. 
        The visualization results of the feature distribution are presented in \Cref{fig:PlotGenImageFeatures}.
        In the quantitative evaluation using IQA metrics, the mean and standard deviation were calculated over 10 verification trials considering the potential for bias in the 50 images for each dataset.

        The visualization results confirmed that the feature distributions of the daytime and night-time images formed two distinct clusters. 
        In particular, the generated daytime image datasets (G3, G4) showed a distribution close to that of the real daytime image datasets (G1, G5) and a different feature distribution from that of the real nighttime image dataset (G2). 

        These results suggest that the generated images possess features similar to those of real daytime images, thereby indicating the potential for domain adaptation through generative data augmentation using D4.

        Moreover, the daytime generated image dataset (G4) accepted by DreamSim exhibits a feature distribution close to that of the real daytime image datasets G1 and G5.
        However, the daytime generated image dataset (G3), which was not from reject DreamSim, showed a somewhat different feature distribution from the real daytime image datasets G1 and G5.
        This suggests that DreamSim's automatic selection mechanism can ensure a certain level of quality in the generated images.
        
        In addition, the daytime generated image dataset (G4) accepted by DreamSim exhibited a feature distribution similar to that of the real daytime image datasets G1 and G5. 
        However, the daytime generated image dataset (G3) rejected by DreamSim showed a somewhat different feature distribution compared to the real daytime image datasets G1 and G5.
        This indicates that the automatic selection mechanism of DreamSim has the potential to ensure a certain quality of generated images.
        However, the feature distribution of the daytime generated image dataset (G4) accepted by DreamSim is observed to be between the distributions of the real image datasets G1 and G5 and the daytime generated image dataset G3 rejected by DreamSim.
        Moreover, in the quantitative evaluation against the daytime real image dataset (G1) for testing, the daytime generated image dataset (G4) accepted by DreamSim showed a slight difference with an FID of 142.09 and a KID of 0.0636 despite the other daytime real image dataset (G5) having an FID of 107.26 and a KID of 0.0057.
        Therefore, the D4 image generation model demonstrated a certain level of practicality; however, it achieved a different quality than actual daytime images.
        
        In our experiments, we confirmed that the detection accuracy did not improve as expected in datasets augmented with more than 500 images generated by D4. 
        This analysis suggests that the generated images have ``image features that differ from the real images'' and ``object annotation mismatch.''
        Moreover, despite utilizing an automatic selection mechanism for images generated using DreamSim, we identified the possibility that its accuracy needs to be improved.
        Therefore, these results imply that stricter control over the quality of the generated images is necessary, thereby suggesting room for improvement in the image-generation process.

        \begin{table}[H]
            \caption{\textcolor{Highlight}{\textbf{Quantitative evaluation of the generated images by IQA metrics}}}
            \includegraphics[draft=false,width=0.5\textwidth]{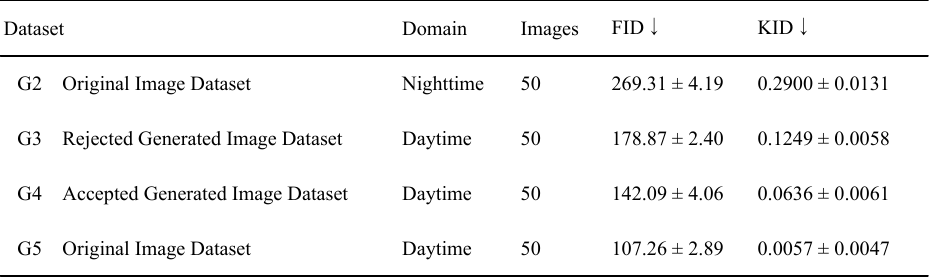}
            \label{table:GenImageIQAmetrics}
        \end{table}

%
\section{Ablation Study}\label{Ablation_Study}
    \subsection{Quality Importance of Generated Images}
        The experimental results confirmed that there was a slight discrepancy in the feature distribution between the generated and real images. 
        Based on these findings, we prepared training datasets of varying qualities for an equal number of generated images. 
        Then, we compared and analyzed the impact on the accuracy of the detection model. 
        The experiment was conducted from two perspectives: (1) The effect of ablating the automatic selection mechanism of the generated images using DreamSim on the congruence of image features was examined. 
        (2) For the congruence of annotations, the impact of manually selected generated images based on human perceptual judgment, considering annotation information, was examined qualitatively for both high- and low-quality generated images. 
        Through this study, the effect of the image quality generated by D4 on the performance of the detection model was analyzed.
    
        As an initial dataset, a set of 100 object-annotated images captured at night was prepared, similar to the experiment described in Section A.
        The dataset was divided into two sets, each containing 50 images: a training set (normal training dataset (a)) and a validation set (normal validation dataset). 
        A style transfer dataset was also prepared using the model trained in stage 1 (Transferred Train Dataset (b)). 
        For comparison, four types of datasets were prepared: two datasets with and without DreamSim's automatic selection mechanism (DreamSim select (c) and (d)) and two datasets curated manually based on human perceptual judgment, one containing only high-quality images (Manual Good-Image select (e)) and the other containing only low-quality images (Manual Bad-Image select (f)).
        The datasets based on human perceptual judgment were selected by focusing on two criteria: the ``alignment between the detected object and annotation'' and the ``visual quality of the image.''

        We conducted comparative studies using the YOLOv8 model for both BBox and keypoint detection. 
        The dataset for validation, the normal validation-dataset, was used during the learning of the normal training dataset (a).
        For learning with other datasets, a transferred validation dataset which ControlNet converted to a learned dataset in stage 1, was used.

        \Cref{table:GenImageSelectIQAmetrics} shows the experimental results.
        For each dataset and model, training was conducted more than five times with different seed values, and the averages and standard deviations of the evaluation metrics were calculated. 
        The experimental results revealed the following two trends:

        \begin{table}[H]
            \caption{\textcolor{Highlight}{\textbf{Accuracy of object detection models for different quality of generated images}}}
            \includegraphics[draft=false,width=1.0\textwidth]{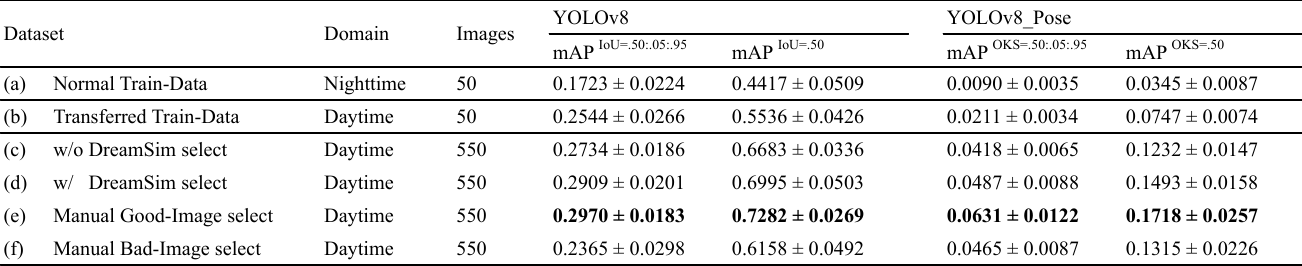}
            \label{table:GenImageSelectIQAmetrics}
        \end{table}

        (1) We discuss the automatic selection mechanism for generated images using DreamSim. 
        The experimental results indicate that dataset (d) with DreamSim-based automatic selection showed improvements in all evaluation metrics for both the BBox and keypoint detection models, compared to that for the dataset (c) without such selection. 

        For the BBox detection model, there was a maximum improvement of 3.12\% in mAP50. 
        For the keypoint detection model, a maximum improvement of 2.61\% was observed in mAP50. 
        These results suggest that the data selected using DreamSim contribute to the performance enhancement of the models.

        (2) We discuss the results of manually selecting the generated images. 
        From the experimental results, dataset (e) with manually selected high-quality generated images showed the highest scores for all evaluation metrics. 
        Compared with the standard dataset (a), the object detection model's mAP50 improved by 28.65\% and that of the keypoint detection model's mAP50 increased by 13.73\%, thereby demonstrating the highest accuracy in both experimental results. 
        Moreover, when compared with dataset (f) containing manually selected low-quality generated images, a significant difference in the evaluation scores was observed, indicating that the quality of the generated images significantly affected the learning of the detection models.

        Thus, the quality of the images generated at D4 is an extremely important factor that significantly influences the training of the detection model. 
        Therefore, D4 requires more accurate image generation models and stricter quality control of the generated images.
        
    \subsection{About Prompt Engineering}
        Prompt engineering is the process of designing optimized input sentences (prompts) for pre-trained large language models such as GPT-3 to generate high-accuracy outputs \citep{PromptEngineering_1_2023}\citep{PromptEngineering_2_2023}.
        The improvement in the accuracy by prompt engineering has been confirmed in CLIP integrated into the architecture of the image generation model StableDiffusion, and its importance is increasing \citep{PromptEngineering_CLIP_2022}. 
        Therefore, we analyzed the effect of prompt engineering on the quality of generated images using the ControlNet model. 
        Through this analysis, we confirmed the effect of applying prompt engineering to improve the quality of the image generation process on D4.

\begin{figure}[H]
            \includegraphics[width=1.0\textwidth]{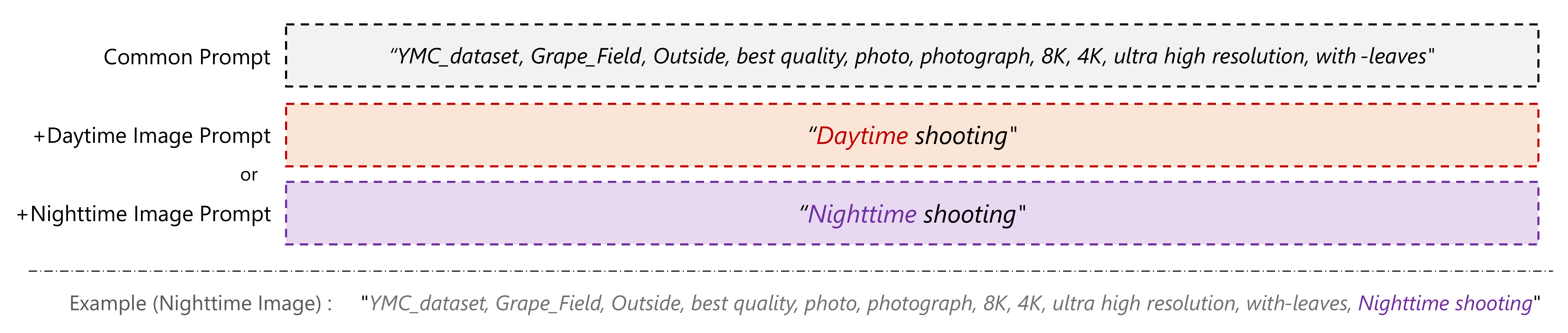}
            \caption{\textcolor{Highlight}{\textbf{Prompt used in this experiment(without prompt engineering)}}}
            \label{fig:Prompts_wo_PE}
        \end{figure}

        The experiment analyzed the impact of changing the prompts used for the pre-training of ControlNet on the quality of the generated images. 
        For the analysis, qualitative evaluation based on human perceptual judgment and quantitative assessment using IQA metrics \citep{IQA_2004} were conducted. 
        In the pre-training of ControlNet, the prompts shown in \Cref{fig:Prompts_w_PE} of \Cref{Experiments_Prompts} were used as prompts with prompt engineering applications, and those shown in \Cref{fig:Prompts_wo_PE} were prepared as prompts without prompt engineering. 
        Moreover, the training utilized the same parameters as in the experiments described in \Cref{Experiments_and_Results}, and the pre-training model selection was based on the best model in the LPIPS metric. 
        A total of 50 images were generated and evaluated for each model. 
        Five types of IQA metrics were employed: distribution-based (DB-IQA) metrics, FID and KID, and full-reference (FR-IQA) metrics, SSIM, LPIPS, and DreamSim.

\begin{figure}[H]
            \includegraphics[width=1.0\textwidth]{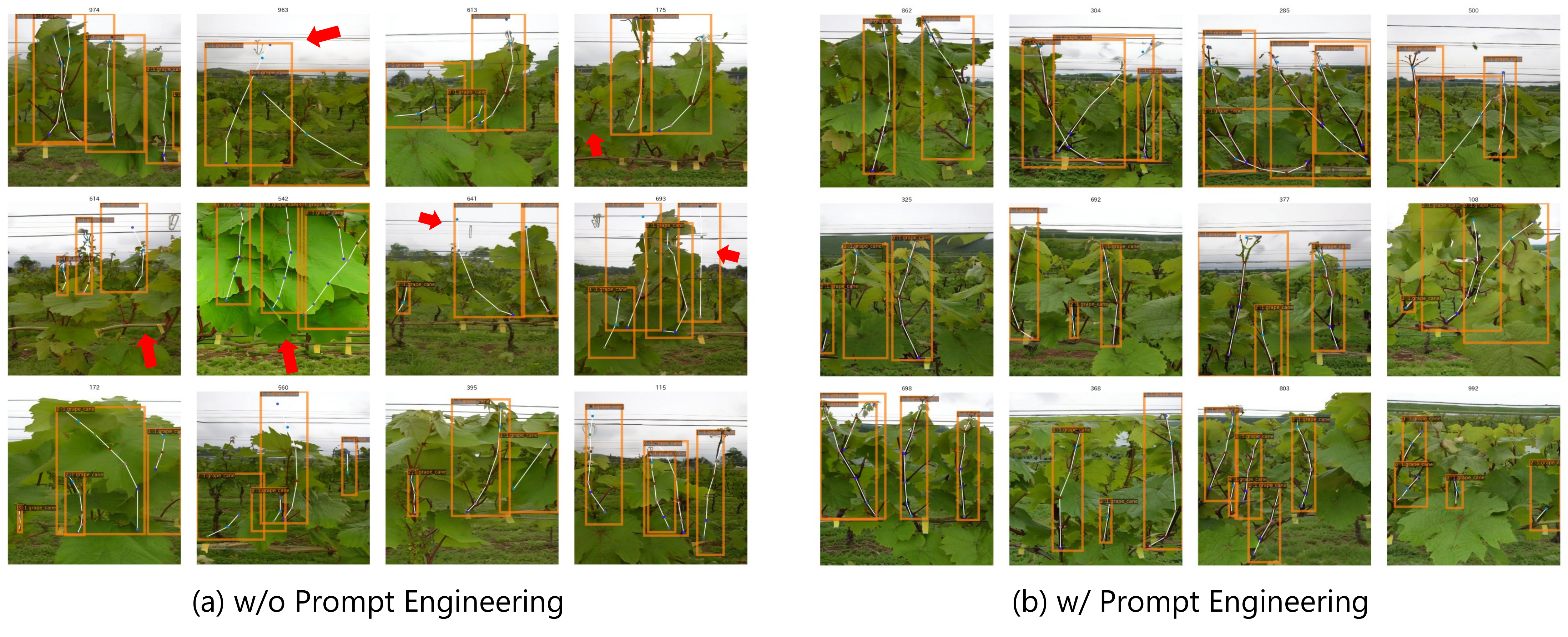}
            \caption{\textcolor{Highlight}{\textbf{Qualitative evaluation based on human perceptual judgment in prompt engineering}}}
            \label{fig:PromptEngineeringQualitative}
        \end{figure}

        We first present the results of a qualitative evaluation based on human perceptual judgment in \Cref{fig:PromptEngineeringQualitative}.
        In models without prompt engineering, inconsistencies were observed between the annotation information and the detected objects, and a significant degradation in image quality were observed.  
        However, in models with prompt engineering, an improvement in the alignment between the annotation information and detected objects was confirmed, and high-quality images were generated. 
        Next, we present the results of the quantitative evaluation based on IQA metrics in \Cref{table:PE_IQAmetrics}.
        For the DB-IQA metrics, improvements in both FID and KID scores were observed, indicating that prompt engineering leads to generated images that are statistically more similar to real images.  
        The adoption of prompt engineering enables the generation of realistic images and faithfully mimics the characteristics of actual image datasets.
        Further, for the FR-IQA metrics, the adoption of prompt engineering achieved quality improvements in all evaluation metrics for the generated images.
        An increase in SSIM indicates an improvement in structural alignment. Meanwhile, the decrease in the LPIPS and DreamSim values suggests that the generated images are more aligned with human perception and that realism has been enhanced.

        These results consistently confirm that the application of prompt engineering improves the quality of image generation models in terms of perceptual similarity, structural alignment, and realism.
        Prompt engineering plays an important role in improving the quality of the generated images.

        \begin{table}[H]
            \caption{\textcolor{Highlight}{\textbf{Quantitative evaluation based on using IQA metrics in prompt engineering}}}
            \includegraphics[draft=false,width=0.5\textwidth]{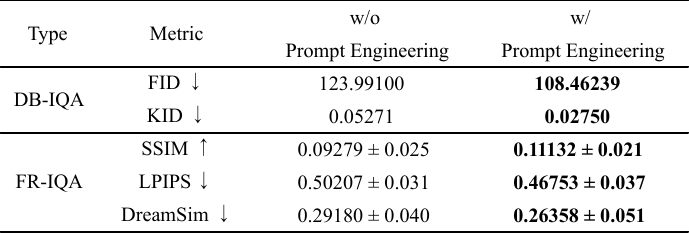}
            \label{table:PE_IQAmetrics}
        \end{table}

%

\section{Conclusion}\label{Conclusion} 
    We proposed a novel generative data augmentation method, D4, using a text-guided diffusion model to automate plant phenotyping in agricultural fields. 
    D4 was designed to address two main challenges: the difficulty of annotation in object-recognition tasks and the diversity of domains. 
    Using D4, it is possible to generate new training data with different domains using a small amount of training data, and this is expected to improve the generality and accuracy of object recognition models. 

    In the experiments, based on a dataset for estimating internode distances in wine grape cultivation, we analyzed the detection accuracy using D4 for BBox detection and keypoint detection in object detection tasks.
    The experimental results confirmed that D4 improves the detection accuracy and demonstrates the effectiveness of domain adaptation. 
    However, it also identified areas of improvement on D4, which concerns the degree of agreement with the annotations. 
    Further, D4 uses an automatic selection mechanism with DreamSim to maintain the quality of the generated images at a certain level, which improves the accuracy of the method. 
    However, a detailed analysis confirmed that further improvements in the quality of the generated images are required. 
    An automatic selection mechanism using DreamSim is required in terms of the degree of agreement between the annotation information and detected objects. 
    Therefore, further improvements are required to increase the accuracy of the image generation model and evaluate the quality of the generated images.
    
    Although many challenges remain, we are confident that the proposed method will contribute to solving the problem of insufficient training datasets in agriculture. 
    We aim to establish more powerful generative data augmentation techniques and improve the accuracy through further analysis and improvements in D4. 
    In addition, we will further investigate the versatility and effectiveness of this method in more complex domain-adaptation tasks by exploring the potential for applications beyond agriculture, such as in the medical and industrial fields.


\section*{CRediT authorship contribution statement}
    Kentaro Hirahara: Conceptualization, Data curation, Formal analysis, Investigation, Methodology, Validation, Visualization, Writing - original draft. 
    Chikahito Nakane: Data curation, Formal analysis. 
    Hajime Ebisawa  : Data curation, Formal analysis. 
    Tsuyosi Kuroda  : Writing - reviewing and editing. 
    Tomoyoshi Utsumi: Investigation, Methodology. 
    Yohei Iwaki     : Investigation, Methodology. 
    Yuichiro Nomura : Writing - reviewing and editing. 
    Makoto Koike    : Writing - reviewing and editing. 
    Hiroshi Mineno  : Project administration, Supervision, Writing - reviewing, and editing.

\section*{Declaration of competing interest}
    The authors declare that they have no competing financial interests or personal relationships that may have influenced the work reported in this study.

\section*{Acknowledgments}
    This research was partially supported by the Japan Science and Technology Agency (JST), FOREST Grant Number JPMJFR201B, and the Research Institute of Green Science and Technology Fund for Research Project Support (2023-RIGST-23205) of the National University Corporation, Shizuoka University. 
    The authors gratefully acknowledge the assistance provided by Naka-izu Winery Hills and Yamaha Motor Co., Ltd. for providing the environment for data acquisition.
    We also express our greatest appreciation to Mr. Bill Coy and Mr. Raul Saldivar from the Yamaha Motor Corporation Unites States (YMUS) for their useful discussions on viticulture and data collection.

\renewcommand\refname{References}
\begin{footnotesize}
\bibliographystyle{elsarticle-num-names} 
\textnormal{\bibliography{References.bib}}
\end{footnotesize}
\newpage

\newpage 
\appendix
\appendixstyle

\section*{Appendix}

\renewcommand{\thefigure}{\Alph{subsection}\arabic{figure}}
\renewcommand{\thetable}{\Alph{subsection}\arabic{table}}
\setcounter{table}{0}
\setcounter{figure}{0}
\subsection{Quality Analysis Method for Generated Images}
        \subsubsection{Overview}
\begin{figure}[H]
                \includegraphics[width=0.7\textwidth]{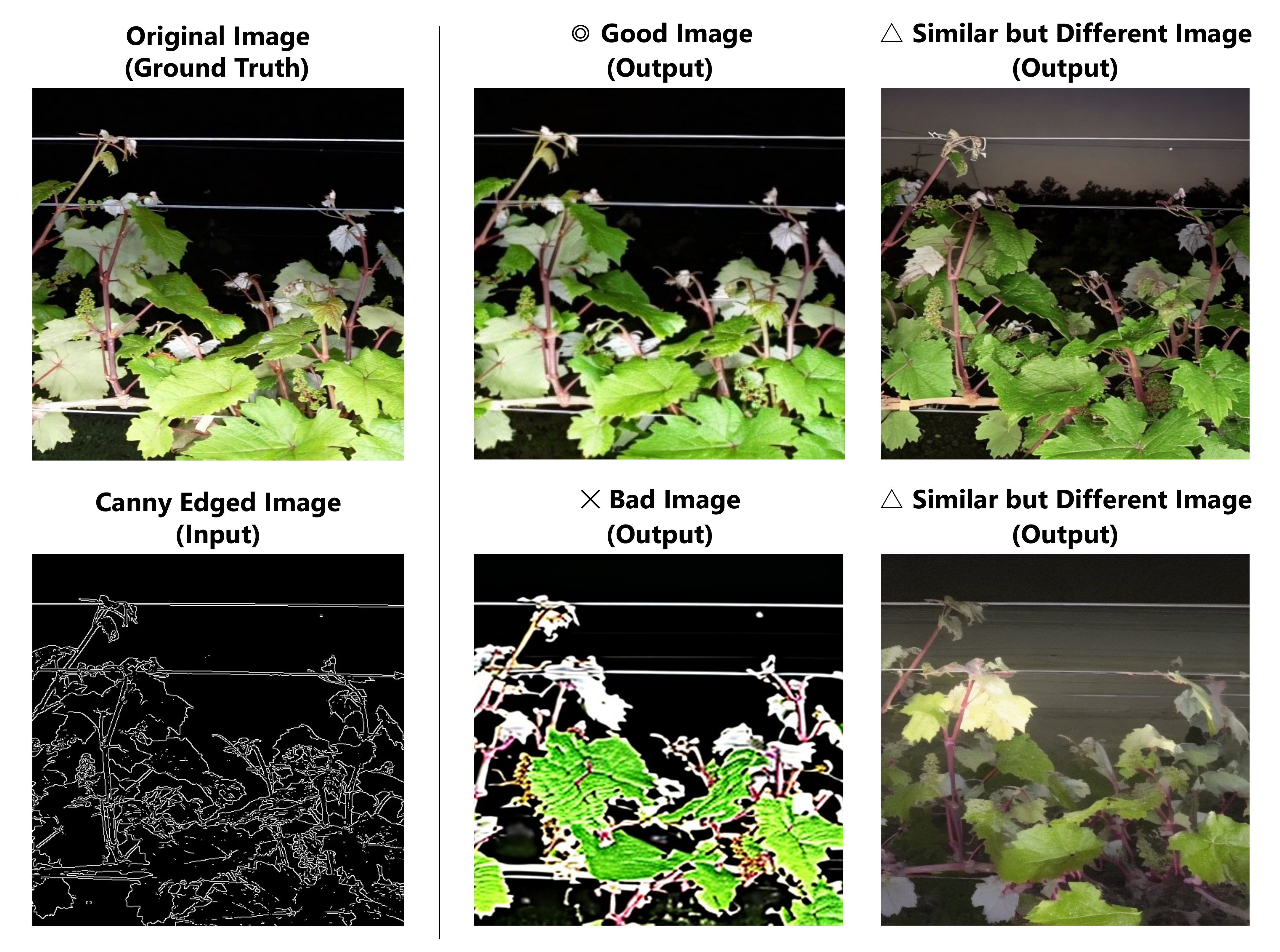}
                \caption{\textcolor{Highlight}{\textbf{Examples of the generated images of different quality}}}
                \label{fig:GenImageQualityExample}
            \end{figure}

            In this study, training a text-guided diffusion model with high generation quality was essential. 
            As shown in \Cref{fig:GenImageQualityExample}, the quality of the generated images varies with each training step, and in some cases, it is clear that ``good images'' and ``bad images'' are generated compared to the original image, while in other cases ``similar but different images'' are generated. 
            Therefore, the optimal learning model was selected by verifying the quality of the generated images at each step using the IQA metric. 
            D4 assumes that the text-guided diffusion model uses the architecture of the existing foundation models without modification. 
            Therefore, the quality of the validation images was quantified using the IQA metric without changing the loss function of the existing model, and an optimal learning model for the text-guided diffusion model was selected. 
            In addition, the approach to the IQA metric was changed based on the learning objectives when selecting the optimal learning model because the learning objectives are different between Stages 1 and 2.

        \subsubsection{Quality Analysis Method at Stage 1 Pre-Train}
            The primary objective of stage 1 learning is to reconstruct the original image from Canny edge images with high accuracy. 
            Images were generated at each learning step of the text-guided diffusion model and the similarity between the original and generated images was verified using the IQA index to select the optimal learning model for stage 1. 
            The IQA index used either a FR or a NR approach. 
            However, DB IQA indices were not used in stage 1 evaluation because the focus is solely on the similarity between images. 
            The model that achieved the highest IQA index score in each learning step was selected as the optimal learning model.

        \subsubsection{Quality Analysis Method at Stage 2 Pre-Train}
            The primary objective of stage 2 learning is to generate images that most closely resemble the image features of real images, where the annotation information matches the detected objects from the annotation-plotted images.
            Therefore, similar to that in stage 1, the images were generated at each learning step of the text-guided diffusion model, and the quality of the generated images was verified using the IQA metric. 
            However, the learning of stage 2 is different from the objective of generating images with the same composition as the original images, and therefore, there is need for a different method to verify the generated images compared with that of stage 1.
            A quality assessment focusing only on annotated parts was conducted in addition to evaluating the quality of the entire generated image. 
            For the IQA metric, either the FR-type or DB-type approach was used. 
            However, in the evaluation of stage 2, the NR-type IQA metric is not used because it required reference images. 
            The model that achieved the best score in the IQA metric at each learning step was selected as the optimal learning model.
            
\begin{figure}[H]
                \includegraphics[width=0.7\textwidth]{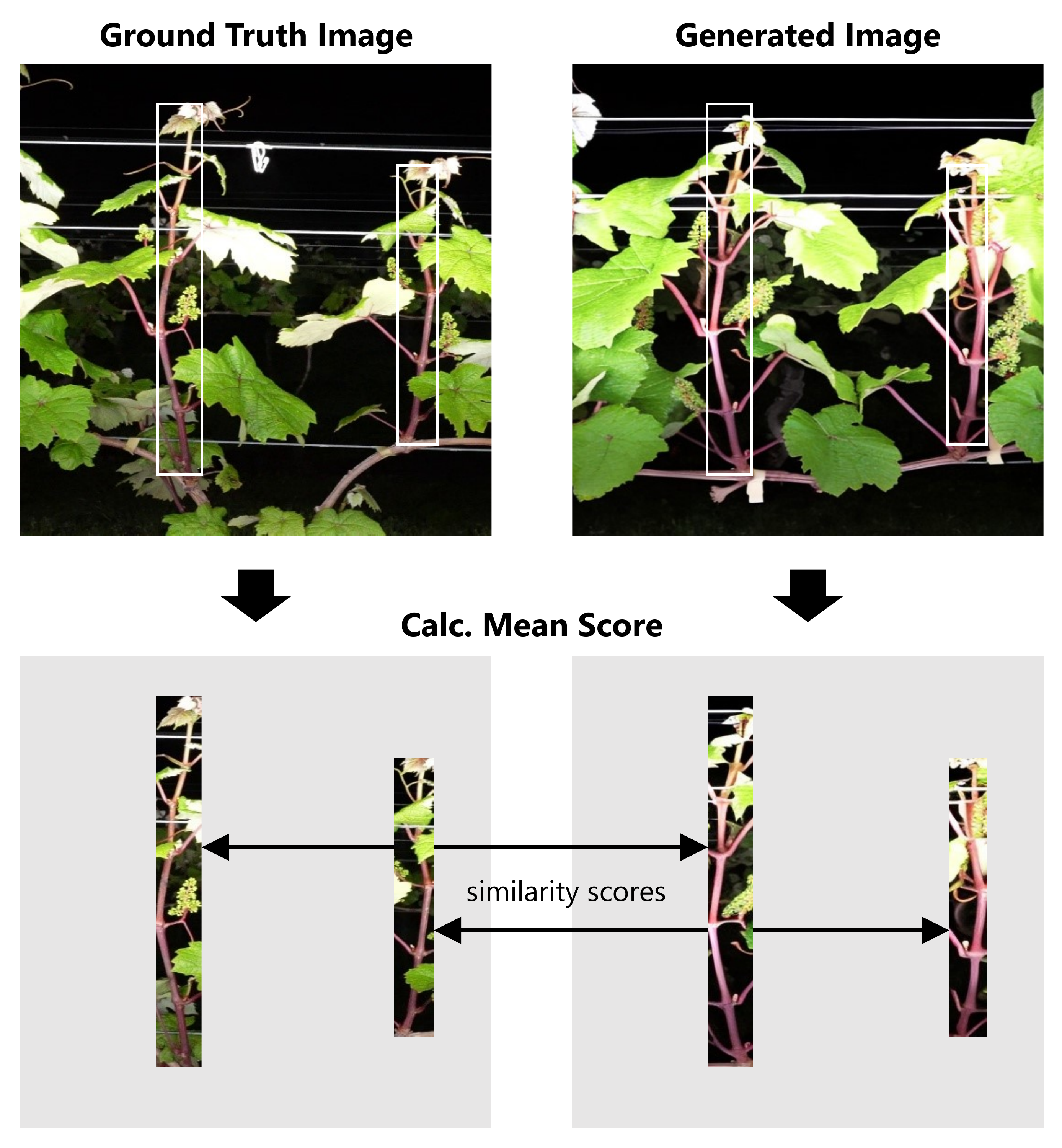}
                \caption{\textcolor{Highlight}{\textbf{Calculation method of evaluation values when using the FR-IQA metrics in Stage 2}}}
                \label{fig:Stage2_FR_IQA}
            \end{figure}

            When the FR-type approach was used as the IQA metric, as shown in \Cref{fig:Stage2_FR_IQA}, the BBox rectangles of the detected target objects were compared.
            This is because the images generated by the stage 2 model must match the annotations. 
            An evaluation value was calculated for each included BBox rectangle, and the average of the evaluation values was determined.

\renewcommand{\thefigure}{\Alph{subsection}\arabic{figure}}
\renewcommand{\thetable}{\Alph{subsection}\arabic{table}}
\setcounter{table}{0}
\setcounter{figure}{0}
\subsection{Pre-Train on ControlNet}
    \subsubsection{Stage 1 Pre-Train}\label{ControlNet_Stage1_Appendix}
        In this section, we detail the pre-training of ControlNet performed in stage 1 of the experiments in this thesis.  
        Verification was performed by generating three images for each of the six predetermined validation images, to obtain a total of 18 images.  
        Model selection was performed using FR-IQA and NR-IQA metrics on the 18 generated images.

        In this study, we implemented and evaluated 16 IQA metrics: 
        Twelve FR–IQA metrics 
            (
                PSNR, 
                MS-SSIM\citep{MS_SSIM_2003}, 
                LPIPS\citep{LPIPS_2018}, 
                the EMD\citep{EMD_2021}, 
                VIF\citep{VIF_2006}, 
                DSS\citep{DSS_2015}, 
                MS-GMSD ~ \citep{MS_GMSD_2017} 
                MDSI\citep{MDSI_2016} 
                VSI\citep{VSI_2014} 
                AHIQ\citep{AHIQ_2022}, 
                TOPIQ ~ \citep{TOPIQ_2023} 
                DreamSim\citep{DreamSim_2023}
            )
        and the four types of NR-IQA metrics
            (
                TV, 
                BRISQUE\citep{BRISQUE_2012}, 
                MANIQA ~ \citep{MANIQA_2022} 
                TOPIQ\citep{TOPIQ_2023}
            ).

\begin{figure}[H]
            \includegraphics[width=1.0\textwidth]{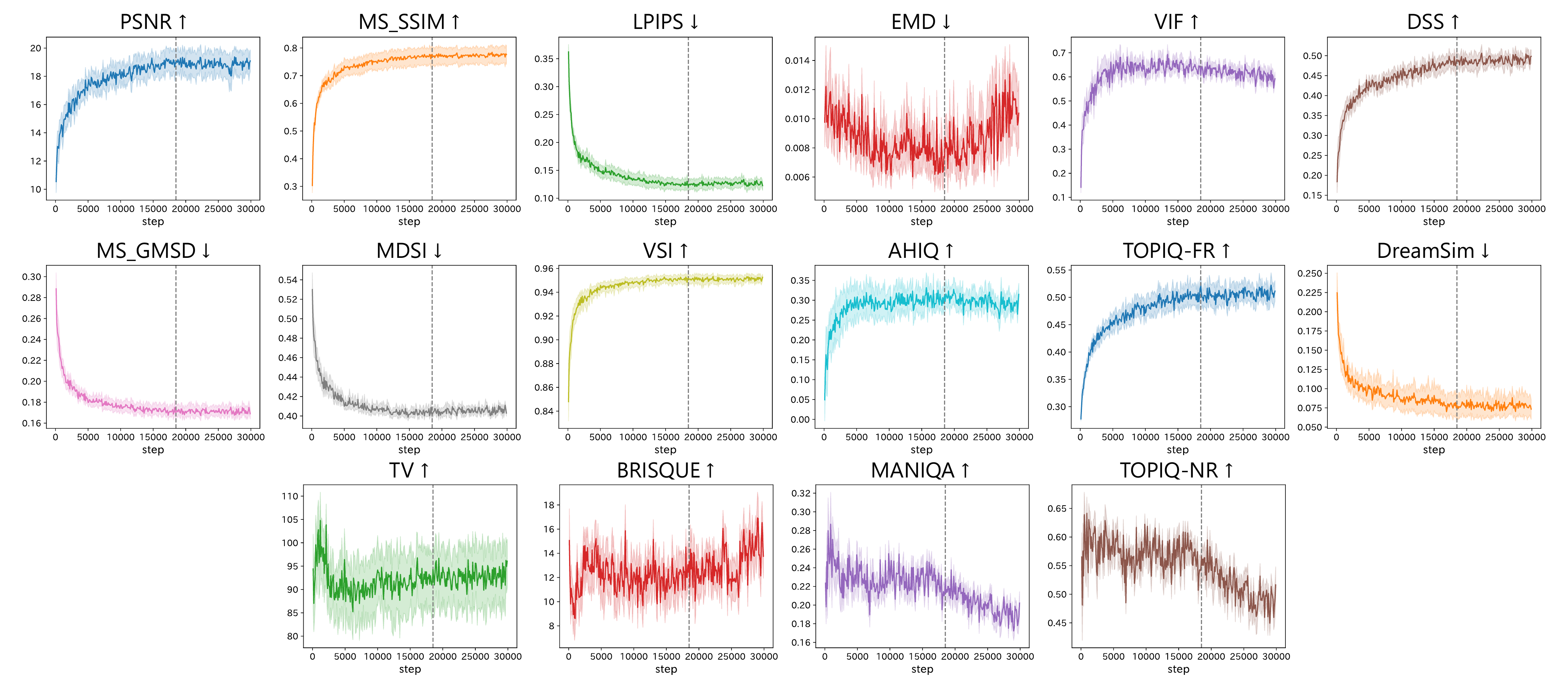}
            \caption{\textcolor{Highlight}{\textbf{T ransition of each IQA metric in the learning process of ControlNet in stage 1}}}
            \label{fig:Stage1_IQA_metrics}
        \end{figure}

        \Cref{fig:Stage1_IQA_metrics} shows the calculation results of each IQA metric during the training process of the stage 1 ControlNet.  
        The figure shows a vertical guideline corresponding to the 18500th step evaluated in the experiment (the best value in the LPIPS).  
        For the FR-IQA metric, we quantitatively confirmed that the quality of the generated images improved as the training progressed.
        However, it is not easy to accurately assess the quality of the generated images using the NR-IQA metric. 
        This suggests that more work is required to evaluate the quality of the generated images accurately because NR-IQA metrics, which do not require reference images (images for comparison), are evaluated based on superficial features such as noise and distortion in the images.
        In addition, fine-tuning was performed using the foundation model of ControlNet. 
        Consequently, a certain level of quality was guaranteed from the beginning of the training, and no significant difference was found in the quantitative values obtained by the NR-IQA metric, which led to such outcomes.

\begin{figure}[H]
            \includegraphics[width=1.0\textwidth]{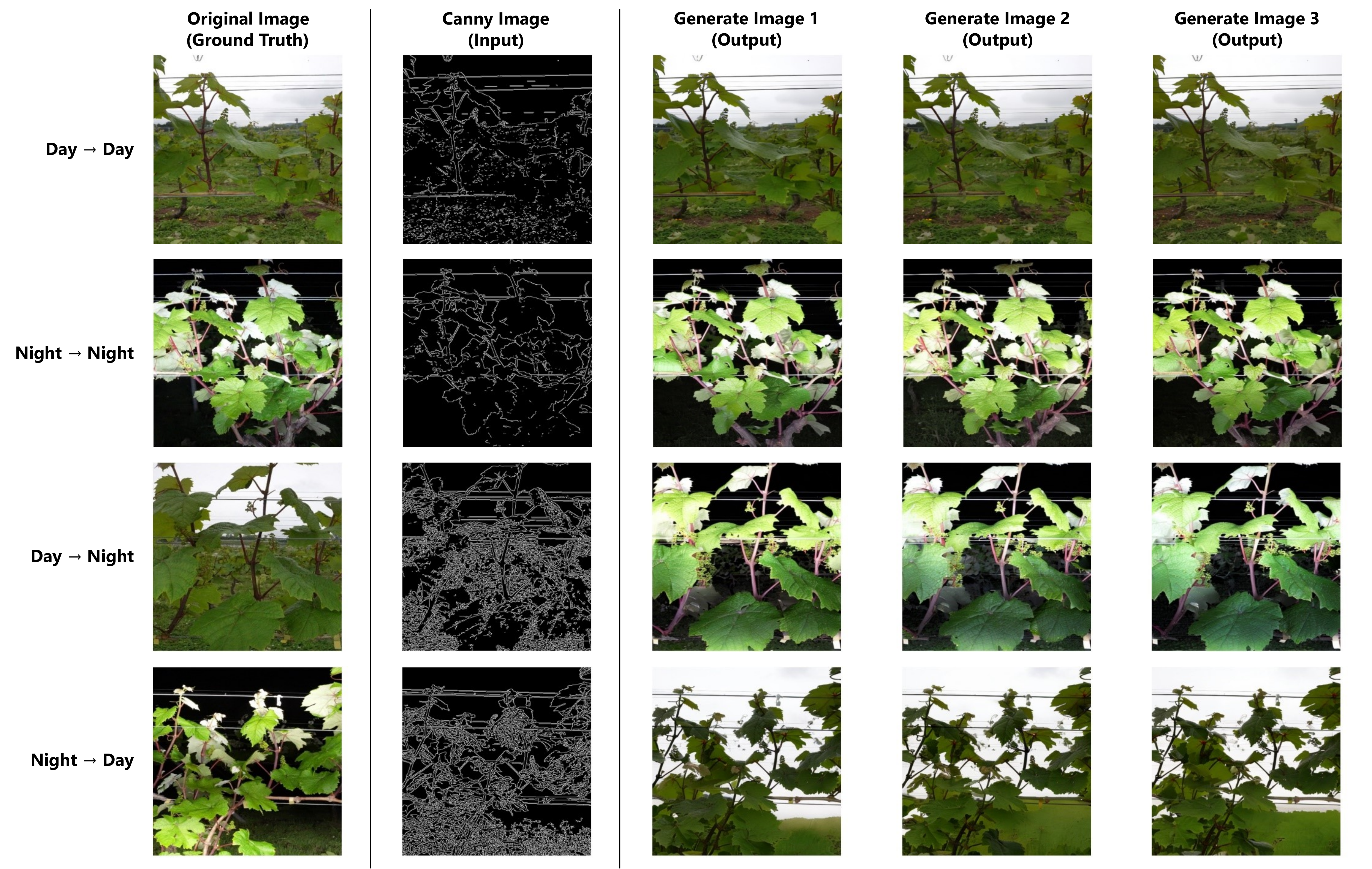}
            \caption{\textcolor{Highlight}{\textbf{Results of the ControlNet model generation for stage 1}}}
            \label{fig:PreTrainResult_Stage1}
        \end{figure}

        \Cref{fig:PreTrainResult_Stage1} shows the results of image generation using the ControlNet model selected at the 18500th step of stage 1 training. 
        In the qualitative evaluation, we confirmed the generation of images with high similarity to the original image when images were generated with the same prompt. 
        In addition, we observed the generation of similar images, albeit with some unnaturalness, when generating images with a different prompt than the original image. 
        This unnaturalness is thought to be influenced by background noise in the Canny border images of the daytime images.  
        When converting daytime images to Canny edges, noise that is not observed in nighttime images may be introduced because of differences in contrast and reflection from background objects. 
        Therefore, the background information in the generated images became unnatural when the Canny edge images extracted from nighttime images were input.

    \subsubsection{Stage 2 Pre-Train}\label{ControlNet_Stage2_Appendix}
        In this section, we describe the details of the pre-training of ControlNet conducted in stage 2 of the experiments in this study. 
        For validation, we used 50 object-annotated images and compared the generated images with the original images for a total of 50 images. 
        The IQA evaluation metrics used were the DB-IQA metrics, FID, and KID. For the FR-IQA metrics, three types of metrics–SSIM, LPIPS, and DreamSim, were calculated based on the results in stage 1.

\begin{figure}[H]
            \includegraphics[width=0.7\textwidth]{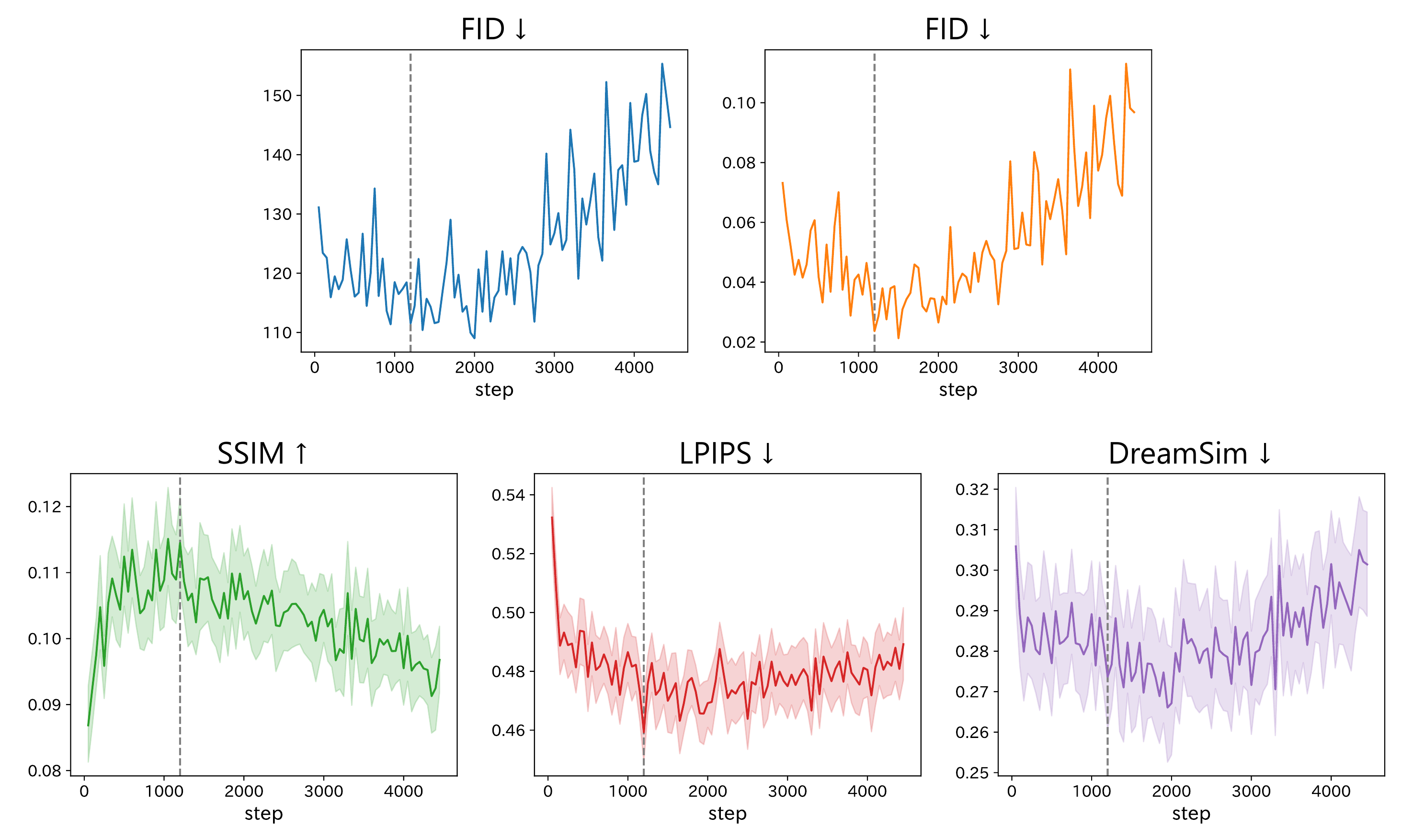}
            \caption{\textcolor{Highlight}{\textbf{Transition of each IQA metric in the learning process of the ControlNet in stage 2}}}
            \label{fig:Stage2_IQA_metrics}
        \end{figure}

        \Cref{fig:Stage2_IQA_metrics} shows the calculation results of each IQA metric during the training process of the stage 2 ControlNet.  
        The figure shows a vertical guideline corresponding to the 1200th step evaluated in the experiment (the best value in the LPIPS).
        For both the DB-IQA and FR-IQA metrics, the quality of the generated images improved from approximately 1000 to 1500 steps, and a decline in quality was observed after 1500 steps.  
        This is attributed to the early convergence of learning caused by the extremely small number of training samples (100). 
        Therefore, the quality of the generated images at each training step in stage 2 learning varied significantly, and careful model selection was necessary.

\renewcommand{\thefigure}{\Alph{subsection}\arabic{figure}}
\renewcommand{\thetable}{\Alph{subsection}\arabic{table}}


\end{document}